\documentclass[11pt,letterpaper,twocolumn,teaser]{planstyle}
\usepackage{microtype}           % Microtypographic refinements
\usepackage{nicefrac}            % Compact symbols for 1/2, etc.
\usepackage{xspace}              % Smart spacing after macros
\usepackage{anyfontsize}
\usepackage{amsmath, amssymb, amsfonts, amsthm, mathtools,mathrsfs}
\usepackage{dsfont}              % Double-stroke font
\usepackage{fdsymbol}            % Additional math symbols
\usepackage[ruled,vlined]{algorithm2e}

\usepackage{booktabs}            % Professional-quality tables
\usepackage{nicematrix}         % Even better sometimes
\usepackage{multirow}
\usepackage{makecell}
\usepackage{bigstrut}
\usepackage{enumitem}            % Customized lists
\setitemize{label=\textbullet, leftmargin=*, nolistsep}
\usepackage{graphicx}
\usepackage{adjustbox}           % Flexible box for figures/tables
\usepackage{float}               % Precise float placement
\usepackage{wrapfig}             % Wrapping figures
\usepackage{subcaption}          % Recommended over subfig
\expandafter\def\csname ver@subfig.sty\endcsname{}  % Avoid conflict with subcaption
\usepackage{fontawesome5}        % Icons
\usepackage{pifont}              % Dingbats
\usepackage{hyperref}            % Hyperlinks
\hypersetup{colorlinks=true, citecolor=LightBlue}
\usepackage[all]{hypcap}         % Fixes links to floats
\usepackage{cleveref}            % Smart cross-referencing
\usepackage{url}                 % For typesetting URLs
\usepackage[normalem]{ulem}      % Underline, strikeout, etc.
\usepackage{lipsum}              % Dummy text
\usepackage{rotating}            % Rotated objects
\usepackage{stackengine}         % Stack text and symbols
\usepackage{caption}             % Caption customization
\usepackage{listings}            % Code listings
\usepackage{soul}                % Undeline (works with break lines)
\usepackage{gradient-text}       % Gradient colored text
\usepackage{pgfplots}            % latex plots
\pgfplotsset{compat=newest}
\usepackage{pgfplotstable}       % for external data tables
\usepackage{tikz}                % Drawing
\usepackage{svg}                 % Include SVG graphics

\usepackage[comma,numbers,sort]{natbib}

\hypersetup{
    colorlinks = true,
    citecolor = {YaleBlue},
}
\definecolor{demphcolor}{RGB}{125,125,125}             % Color for deemphasis
\newcommand{\ie}{\textit{i.e.},\xspace}      % i.e.
\newcommand{\eg}{\textit{e.g.},\xspace}      % e.g.
\crefname{equation}{Eq.}{Eqs.}
\crefformat{section}{\S#2#1#3}
\crefformat{subsection}{\S#2#1#3}
\crefformat{subsubsection}{\S#2#1#3}
\crefrangeformat{section}{\S\S#3#1#4 to~#5#2#6}
\crefmultiformat{section}{\S\S#2#1#3}{ and~#2#1#3}{, #2#1#3}{ and~#2#1#3}
\newcommand{\cmark}{\textcolor{ForestGreen}{\ding{51}}}  % ✓
\newcommand{\xmark}{\textcolor{red}{\ding{55}}}     % ✗

\newcommand{\scalemath}[2]{\scalebox{#1}{$#2$}}
 \usepackage[dvipsnames]{xcolor}
\newcommand{\modelname}{CoRe3D\xspace}

\newcommand{\modelnamegradient}
{\textbf{\gradientRGB{CoRe3D}{0, 128, 128}{136, 74, 178}}\xspace}

\title{\modelnamegradient: Collaborative Reasoning as a \\Foundation for 3D Intelligence}

\author{
\vspace{-0.2cm}
    \begin{tabular}{c}
    \textbf{Tianjiao Yu, Xinzhuo Li, Yifan Shen, Yuanzhe Liu, Ismini Lourentzou}\\
    {\scriptsize{\tt\{ty41, lourent2\}@illinois.edu}} 
    \end{tabular}
}

\affil{\textcolor{IllinoisBlue}{University of Illinois Urbana-Champaign}}

\correspondingauthor{
$^*$ Preprint. Work in progress.
}
\begin{document}
\setabstractlogo[9mm]{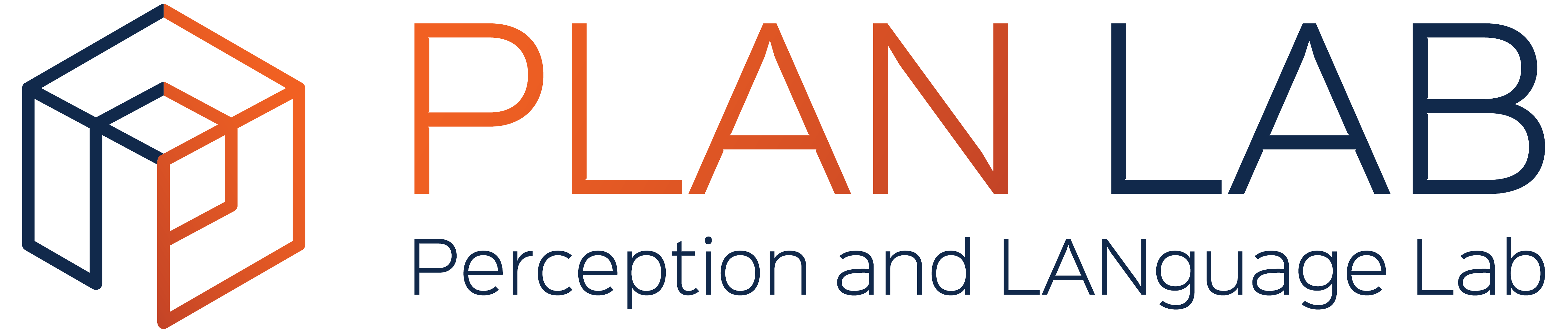} 
\begin{abstract}
Recent advances in large multimodal models suggest that explicit reasoning mechanisms play a critical role in improving model reliability, interpretability, and cross-modal alignment. While such reasoning-centric approaches have been proven effective in language and vision tasks, their extension to 3D remains underdeveloped. 
\modelname introduces a unified 3D understanding and generation reasoning framework that jointly operates over semantic and spatial abstractions, enabling high-level intent inferred from language to directly guide low-level 3D content formation. Central to this design is a spatially grounded reasoning representation that decomposes 3D latent space into localized regions, allowing the model to reason over geometry in a compositional and procedural manner. By tightly coupling semantic chain-of-thought inference with structured spatial reasoning, \modelname produces 3D outputs that exhibit strong local consistency and faithful alignment with linguistic descriptions.  
\vspace{3.5mm}
\\ \url{https://plan-lab.github.io/core3d}
\vspace{-2mm}
\end{abstract}
\maketitle

\section{Introduction}
\label{sec:intro}
Despite rapid progress in 3D generation, most existing methods remain imitation-based, reproducing shapes rather than reasoning about objects \cite{poole2022dreamfusion, yi2024gaussiandreamer}. As a result, they struggle when prompts implicitly specify structure (\eg relations, counts, geometry, or physical contact). In contrast, recent unified vision–language models have started to capture these same signals effectively in 2D settings~\cite{yang2024qwen2,touvron2023llama}.
This progress is largely attributed to the integration of Chain-of-Thought (CoT) reasoning~\cite{wei2022chain}, which, when extended to multimodal LLMs~\cite{chen2025r1v,meng2025mm,r1vl,zhang2024mavis}, improves interpretability and consistency across visual reasoning tasks~\cite{Lu2023MathVistaEM,jiang2025mme}.

\begin{figure*}[!t]
\centering
    \includegraphics[width=0.99\linewidth]{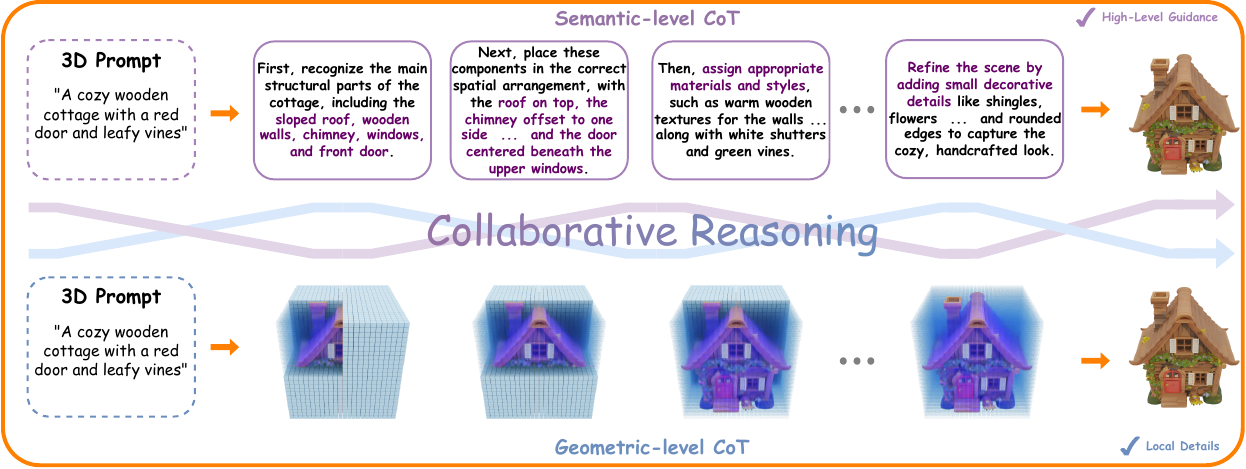}
    \vspace{0.1cm}
    \caption{We introduce \modelnamegradient, a framework that unifies \textcolor{CustomPurple}{Semantic CoT} and octant-based \textcolor{CustomTeal}{Geometric CoT} through collaborative reasoning. By coupling language-grounded reasoning with shape constructions, \modelname enables bidirectional capability in both 3D understanding and generation, allowing the model to interpret objects and construct them within a unified framework.}
    \label{fig:teaser}
\end{figure*}

However, analogous unified reasoning in 3D remains under-explored. Few models can both interpret observations and construct a consistent 3D object representation within a single framework~\cite{ye2025shapellm,wang2024llama}. To advance this frontier, we propose \modelnamegradient, a framework for collaborative reasoning that unifies semantic understanding and geometric generation within a single 3D-LLM. As illustrated in Fig.~\ref{fig:teaser}, \modelname integrates a unified 3D language model with an octant-based 3D VQ-VAE, enabling the model to reason in both language and 3D token space.\looseness-1

At its core, our approach couples a \textcolor{CustomPurple}{Semantic CoT} for high-level textual planning with a novel \textcolor{CustomTeal}{Geometric CoT} for spatial synthesis. The geometric CoT operates autoregressively across octant blocks, addressing the limitations of existing “flat” voxel representations that waste computation on empty space and fail to capture structured spatial dependencies. Unlike part-level representations that require fixed ontologies and suffer from poor generalization across categories \cite{chen2025autopartgen}, or voxel-level representations that remain unstructured and semantically agnostic \cite{qi2024shapellm,xiang2024structured}, our octant-based representation remains ontology-free yet structure-aware.\looseness-1

To jointly refine both reasoning streams, we employ \textbf{Co-GRPO}, a collaborative extension of GRPO~\cite{shao2024deepseekmath} that optimizes \modelname with \emph{multi-critic}, 3D-aware rewards. Concretely, an ensemble of complementary critics balances semantic intent, perceptual quality, text–3D alignment, and physical coherence, providing dense feedback over both the semantic and geometric traces. This design is crucial because it (1) elicits executable plans even when no ``gold'' reasoning supervision exists, (2) enables fine-grained process credit assignment through dense, 3D-specific rewards, and (3) improves training robustness by aggregating complementary signals, reducing sensitivity to any single imperfect evaluator. By jointly rewarding linguistic reasoning and 3D synthesis, Co-GRPO moves toward general 3D intelligence that unifies understanding and generation.\looseness-1

\noindent \textbf{Contributions:} In summary, our contributions are:
\begin{itemize}[itemsep=0.5ex, parsep=0pt, topsep=-2.3pt, leftmargin=0.4cm]
    \item We introduce \modelnamegradient, a collaborative reasoning framework that unifies two complementary reasoning levels, a \textcolor{CustomPurple}{Semantic CoT} for textual planning, with a novel octant-based \textcolor{CustomTeal}{Geometric CoT} that acts as a structure-aware yet ontology-free prior, enabling interpretable progressive construction without category-specific part definitions.

    \item To the best of our knowledge, we are the first to use Co-GRPO to jointly optimize semantic and geometric reasoning in 3D. This approach elicits plans without direct supervision and effectively assigns credit using dense 3D-specific rewards (\eg symmetry, physical coherence) for improved alignment, structure, and robustness.

    \item We further demonstrate that our unified reasoning paradigm is not limited to generation but naturally extends to reciprocal 3D understanding tasks, such as 3D-to-text captioning and reasoning-intensive text-to-3D, highlighting its potential as a scalable foundation for general 3D intelligence.
\end{itemize}

\section{Related Work}
\textbf{3D Generation.} 
Early text-to-3D frameworks~\cite{poole2022dreamfusion,wang2023prolificdreamer,chen2023fantasia3d,lin2023magic3d,raj2023dreambooth3d,sweetdreamer,sun2023dreamcraft3d,chen2024text,sjc,tang2023dreamgaussian,yi2024gaussiandreamer} formulated 3D synthesis as an optimization problem guided by 2D priors through score distillation sampling (SDS). 
While this approach enabled cross-modal 3D generation without paired data, it required long per-instance optimization and often produced view-inconsistent geometry.
Subsequent methods~\cite{wang2023imagedream,shi2023mvdream,wang2023animatabledreamer,ye2024dreamreward,qiu2024richdreamer,chen2024microdreamer} address view inconsistency by enforcing cross-view semantic constraints within diffusion pipelines. 
Other works address the inefficiency of iterative optimization~\cite{long2024wonder3d,zhao2024flexidreamer,liu2023syncdreamer,liu2023zero,shi2023zero123++,weng2023consistent123,liu2023one,wu2024unique3d,chen2024v3d,voleti2024sv3d,ye2024stablenormal,liu2024reconxreconstructscenesparse, yu2025part, gao2024partgs} by first generating consistent 2D renderings then reconstructing the 3D geometry through fast neural reconstruction.

More recently, native 3D diffusion models~\cite{zhao2023michelangelo,wang2023rodin,wu2024direct3d,yang2024hunyuan3d,huang2025spar3d,zhang2024clay,xiang2024structured,chen20243dtopia,li2024craftsman,ye2025hi3dgen} shifted toward generative modeling within latent 3D spaces, employing VAE-based encoders to learn volumetric or implicit shape priors. In contrast to diffusion models, approaches employ vector-quantized autoencoders~\cite{van2017neural}, casting 3D generation as an autoregressive sequence modeling problem~\cite{siddiqui2024meshgpt, chen2024meshanything, weng2024pivotmesh}. 
Later works~\cite{chen2024meshanythingv2,chen2025meshxl,weng2024scaling,tang2024edgerunner,hao2024meshtron,zhao2025deepmesh, wang2024llama} introduce task-specific tokenization schemes that directly encode vertex–face structures, improving geometric fidelity and local continuity.  However, these models still operate as next-token predictors and do not expose an explicit reasoning process. Our method instead pairs an octant-based 3D VQ-VAE with a unified 3D-LLM that performs both semantic and geometric chain-of-thought reasoning, leading to better 3D generation and understanding performance.

\noindent\textbf{Unified Generation \& Understanding.} Recent multimodal LLMs have revealed remarkable capability in jointly processing and generating vision–language content. Early frameworks~\cite{alayrac2022flamingo,bai2023qwen,chen2024internvl} extend LLMs with visual encoders for grounded perception, while more recent systems~\cite{Chameleon_Team_Chameleon_Mixed-Modal_Early-Fusion_2024,liu2024world,wang2024emu3,xie2024show,zhou2024transfusion} integrate text and image generation through learned visual tokenizers and mixed-modality training.

Extending this paradigm to 3D, emerging studies~\cite{hong20233d,xu2024pointllm,qi2024shapellm,xue2023ulip,chen2024ll3da} adapt LLMs for 3D understanding using point-cloud or shape embeddings. 
While effective for perception tasks, these models largely focus on recognition rather than generation. Subsequent efforts~\cite{yin2025shapegpt,wang2024llama,chen2024sar3d,ye2025shapellm,zhu20233d} attempt to unify language and 3D modeling by developing generative LLMs that handle both understanding and generation within a shared representation space. 
More interactive paradigms, such as LL3M~\cite{lu2025ll3mlargelanguage3d} and L3GO~\cite{yamada2024l3golanguageagentschainof3dthoughts}, employ agent-based reasoning to iteratively construct or edit 3D scenes, yet rely on symbolic planning rather than token-level 3D reasoning. In contrast, our approach integrates semantic and geometric reasoning within a unified 3D-LLM. By explicitly modeling the reasoning process across both language and 3D token spaces, \modelname achieves reasoning-aware 3D understanding and generation, bridging the gap between linguistic intent and physically grounded 3D synthesis.
\begin{figure*}[!t]
\centering
    \includegraphics[width=0.85\linewidth]{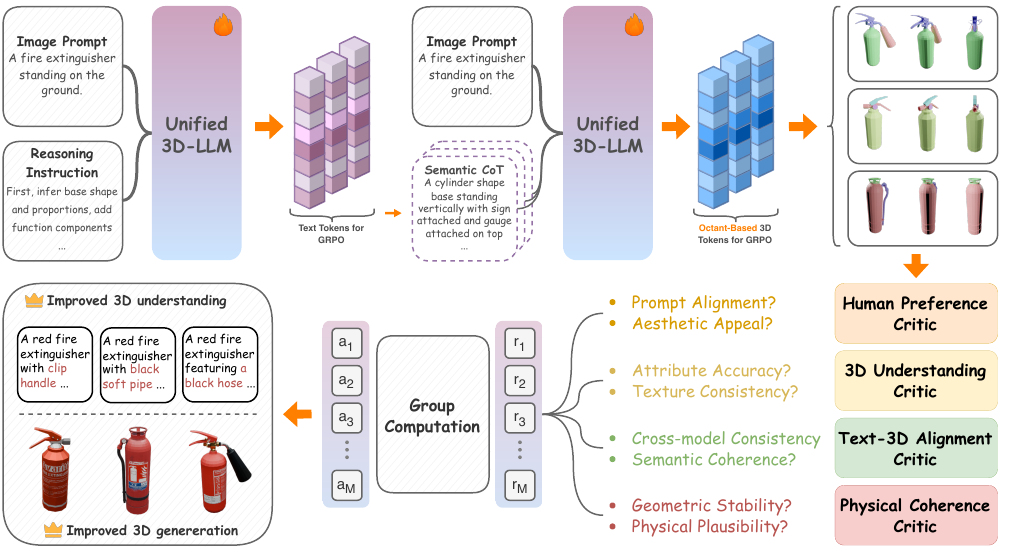}
    \caption{\textbf{\modelnamegradient overview.} Semantic and geometric reasoning tokens are generated by our unified 3D-LLM, and the generated 3D object and corresponding multiviews are evaluated by an ensemble of critics.}
    \label{fig:main_method}
\end{figure*}

\section{Method}
\label{sec:method_main}

\subsection{Semantic and Geometric Representations}

\noindent\textcolor{CustomPurple}{\textbf{Semantic-Level Representation.}}
A core challenge in 3D generation is translating open-ended language into structured reasoning signals that preserve compositional semantics and physical constraints. Directly mapping prompts into latent 3D tokens is under-specified, as language descriptions typically omit precise geometric, relational, and material cues, resulting in generated shapes that capture coarse appearance but fail to recover consistent structure or texture details. To address this gap, we introduce a \textit{semantic CoT} reasoning stage that expands each textual prompt into an explicit structural plan before geometry generation.
Given an input description and an optional reasoning instruction, the unified 3D-LLM first produces a detailed natural language description of the object category, spatial layout, materials, and appearance details. This description serves as an interpretable, text-based scaffold that anchors the subsequent geometric reasoning process. 
Formally, we represent the semantic reasoning trace as a sequence of tokens 
$\mathcal{S}_\text{sem} = [s_1, s_2, \dots, s_N]$, 
conditioned on the input prompt and reasoning instruction. 
These tokens are optimized jointly with 3D generation tokens, enabling semantic intent to directly modulate spatial synthesis during training.\looseness-1

\noindent\textcolor{CustomTeal}{\textbf{Geometric-Level Representation.}} 
We represent each 3D object using a $64^3$ voxel grid, which provides a balanced trade-off between spatial fidelity and computational efficiency. To obtain a compact token representation, a 3D VQ-VAE encoder maps the voxel grid to a $16^3$ latent grid, preserving local geometry and appearance features.  The latent grid is serialized into 4096 latent vectors, each corresponding to a spatial location. To further reduce sequence length, we group every $2{\times}2{\times}2$ neighborhood of latent voxels (eight adjacent cells) by concatenating their channels into a single vector. This operation transforms the 4096 latent vectors (with 8-D channels) into 512 tokens with 64-D channels, where each token represents a local \textit{octant block} within the 3D volume.\looseness-1

A vector-quantization module with an 8192-entry codebook discretizes these octant features, resulting in 512 discrete 3D tokens per object. 
For spatial disambiguation across blocks, we attach a learned absolute position embedding to each octant token, keyed by its $(x_b,y_b,z_b)$ index on the $8{\times}8{\times}8$ block grid (or an equivalent Morton/Z-order code). This embedding is injected post-quantization, so the codebook remains content-centric while the generator remains location-aware.\looseness-1

This octant-based representation naturally supports our \textit{geometric CoT} reasoning: the model iteratively ``thinks'' over octant tokens, refining or sampling candidate completions for each sub-cube.
We denote the sequence of geometric reasoning tokens as 
\begin{equation}
    \setlength{\abovedisplayskip}{6pt}
\setlength{\belowdisplayskip}{6pt}
\mathcal{G}_{\text{geo}} = [g_1, g_2, \dots, g_M],
\end{equation}
where each token $g_i$ corresponds to a discrete octant block produced by the 3D VQ-VAE. 
While the \textcolor{CustomPurple}{semantic CoT} $\mathcal{S}_{\text{sem}}$ expresses the conceptual plan in language space, 
the \textcolor{CustomTeal}{geometric CoT} $\mathcal{G}_{\text{geo}}$ realizes that plan token-by-token in the 3D latent space. 
Together, these two reasoning levels enable controllable and interpretable 3D generation that preserves both global semantics and local geometric precision.

\subsection{Preliminaries}
\label{sec:prelim}
Reinforcement learning has recently become a dominant tool for eliciting reasoning behaviors in large models. A particularly effective variant is Group Relative Policy Optimization (GRPO) \cite{shao2024deepseekmath}, which modifies Proximal Policy Optimization (PPO) by discarding the explicit value function and instead normalizing rewards within a sampled group of trajectories.\looseness-1

Formally, given a prompt–answer pair $(p,a)$, the old policy $\pi_{\theta_{\text{old}}}$ generates a group of $G$ candidate responses $\{o_i\}_{i=1}^G$. Each response is scored by a reward model, yielding $\mathcal{R}_i$. To reduce variance and emphasize relative quality, the advantage of the $i$-th response is defined by standardizing rewards within the group:\looseness-1
\begin{equation}
\setlength{\abovedisplayskip}{6pt}
\setlength{\belowdisplayskip}{6pt}
A_i = \frac{\mathcal{R}_i - \mu(\{\mathcal{R}_j\}_{j=1}^G)}{\sigma(\{\mathcal{R}_j\}_{j=1}^G)},
\end{equation}
where $\mu$ and $\sigma$ denote the mean and standard deviation.  
The learning objective follows the clipped surrogate structure of PPO, but with a direct KL penalty that anchors the updated policy $\pi_\theta$ to a reference distribution $\pi_{\theta_{\text{ref}}}$:
\begin{multline}  
\setlength{\abovedisplayskip}{6pt}   
\setlength{\belowdisplayskip}{6pt} 
\scalemath{0.95}{
\mathcal{J}_{\text{GRPO}}(\theta) =
\mathbb{E}_{\{o_i\}_{i=1}^G} 
\Bigg[ \frac{1}{\sum\limits_{i=1}^G |o_i|} \sum\limits_{i=1}^G \sum\limits_{t=1}^{|o_i|}
\Big( \min \big( r_{i,t}(\theta) A_i, \; 
}\\
\scalemath{0.95}{
\text{clip}(r_{i,t}(\theta), 1-\varepsilon, 1+\varepsilon) A_i \big) 
- \beta D_{\text{KL}}(\pi_\theta \,||\, \pi_{\theta_{\text{ref}}})
\Big) \Bigg],  
}
\label{eq:grpo_objective}
\end{multline}
where the importance ratio at each token step is
\begin{equation}
r_{i,t}(\theta) = \frac{\pi_\theta(o_{i,t}\mid p, o_{i,<t})}{\pi_{\theta_{\text{old}}}(o_{i,t}\mid p, o_{i,<t})}.
\end{equation}
Here, $\varepsilon$ controls the clipping range of the importance ratio, and $\beta$ determines the strength of the KL penalty that keeps the updated policy close to the reference policy $\pi_{\theta_{\text{ref}}}$.\looseness-1

While GRPO has been applied mainly in text reasoning (\eg math or code generation), its principle of \textit{relative quality comparison} remains under-explored in 3D generation. Leveraging this idea, we evaluate groups of rollout trajectories—spanning both semantic plans and geometric refinements—against one another, where their relative ranking provides a stable optimization signal that aligns outputs with semantic intent and geometric plausibility.

\noindent \textbf{Octree-based Autoregressive Model.} Recent 3D transformers such as OctFormer~\cite{wang2023octformer} demonstrate that representing 3D data using an \textit{octree hierarchy} enables efficient global reasoning while preserving local geometric detail. Instead of processing dense voxel grids, an octree partitions the 3D space into hierarchical cubic cells (octants) of adaptive resolution, allocating finer subdivisions in geometrically complex regions and coarser ones in uniform areas. This sparse yet structured representation significantly reduces memory and computation costs compared to dense attention over all voxels.

Formally, an octree representation can be described as a set of hierarchical nodes 
$\mathcal{O} = \{ o_\ell^k \mid \ell \in [0, L), k \in \mathcal{I}_\ell \}$, 
where $\ell$ denotes the level in the tree and $k$ indexes the spatial position at that level. 
Each node $o_\ell^k$ encodes geometric and visual features (\eg occupancy, color, or normal) aggregated from its eight child nodes at level $\ell{+}1$. 
The model applies transformer attention hierarchically across this structure: 
\textit{intra-level attention} aggregates context among nodes within the same resolution, while \textit{inter-level attention} propagates information between parent and child nodes to capture cross-scale dependencies.

This design provides two key advantages: (1) it maintains spatial locality, allowing the model to focus computation on occupied regions, and 
(2) it establishes a natural coarse-to-fine reasoning pathway across the 3D volume. In our work, we adopt a simplified variant of this idea by discretizing the latent 3D volume into uniform $2{\times}2{\times}2$ octant blocks rather than an adaptive octree hierarchy. Each octant token thus serves as a fixed-resolution counterpart to an octree node, preserving the locality of the model while enabling autoregressive reasoning through the geometric CoT process described in the following sections.

\subsection{Collaborative Reasoning}

\noindent The core innovation of our framework lies in the explicit collaboration between \textcolor{CustomPurple}{semantic} and \textcolor{CustomTeal}{geometric} reasoning. While each level can operate independently, their joint optimization leads to stronger, mutually reinforcing behavior. We unify them through \textbf{3D Co-GRPO}, a reinforcement learning framework that refines both reasoning levels using multi-critic 3D-aware rewards, aligning linguistic intent with spatial construction. This results in objects that are semantically faithful, visually compelling, and physically coherent. An overview is shown in~\Cref{fig:main_method}.

Formally, given an input prompt $T_p$ and pre-generated semantic CoT decoded from $\mathcal{S}_{\text{sem}}$, the unified 3D-LLM produces a geometric reasoning sequence $\mathcal{G}_{\text{geo}}\!=\![g_1, g_2, \dots, g_M]$ to synthesize the final 3D object $\hat{O}$. 
The process can be viewed as a sequential reasoning pipeline: 
the semantic trace $\mathcal{S}_{\text{sem}}$ provides global planning cues such as object category, spatial relations, and texture details, while the geometric trace $\mathcal{G}_{\text{geo}}$ progressively realizes those cues within the latent 3D token space. 

We extend the GRPO paradigm into the 3D domain by introducing four complementary \emph{critics}, each providing a scalar reward that captures a distinct dimension of 3D quality. The resulting meshes and multi-view renderings $\{I_i\}$ are evaluated by an ensemble of 3D experts:\looseness-1
\begin{itemize}[itemsep=0.5ex, parsep=0pt, topsep=-2.3pt, leftmargin=0.4em]
    \item \textbf{Human Preference Critic}: evaluates perceptual realism and human aesthetic preference, assessing prompt relevance and overall visual appeal from multi-view renderings~\cite{wu2023human, xu2023imagereward, ye2024dreamreward}.
    \item \textbf{3D Understanding Critic}: verifies attribute- and part-level correctness using 3D-VQA models~\cite{zhu20233d, chen2024ll3da, qi2024shapellm, hong20233d} that query geometric, textural, and symmetry attributes derived from the semantic CoT $\mathcal{S}_{\text{sem}}$.
    \item \textbf{Text–3D Alignment Critic}: measures semantic faithfulness between the textual reasoning trace (prompt and $\mathcal{S}_{\text{sem}}$) and the generated geometry $\mathcal{G}_{\text{geo}}$, using pretrained text–3D embedding models~\cite{xue2024ulip, yin2025shapegpt} to ensure cross-modal coherence and faithful alignment.\looseness-1
    \item \textbf{Physical Coherence Critic}: analytically enforces structural stability and physical plausibility through a differentiable reward composed of three geometry-based terms:
    \(
    R_P = \lambda_1 R_{\text{stab}} + \lambda_2 R_{\text{rig}} + \lambda_3 R_{\text{int}},
    \)
    where $R_{\text{stab}}$ measures global balance of the center of mass, $R_{\text{rig}}$ promotes topologically connected geometry that maintains physical continuity, and $R_{\text{int}}$ penalizes self-intersection between surfaces. This critic is fully geometry-driven and provides a compact measure of physical coherence.\looseness-1
\end{itemize}
Each critic outputs a normalized reward $R_i \in [0,1]$, and the overall GRPO objective aggregates them via
\begin{equation}
R = w_H R_H + w_V R_V + w_X R_X + w_P R_P,
\label{eq:grpo_reward}
\end{equation}
where weights balance human preference, 3D understanding, cross-modal alignment, and physical coherence. The combined reward provides preference signals for policy updates, encouraging the model to improve both reasoning accuracy and geometric fidelity over GRPO iterations.

During training, the model first performs forward generation to produce $\mathcal{S}_{\text{sem}}$ and $\mathcal{G}_{\text{geo}}$ for each prompt $T_p$. Generated results are rendered into multi-view images and passed through the four critics, producing individual rewards $\{R_H, R_V, R_X, R_P\}$. The composite reward in ~\cref{eq:grpo_reward} is then used to compute pairwise preferences among samples and update the model using the GRPO objective defined in ~\cref{eq:grpo_objective}.

\section{Experiments}
\noindent \textbf{Implementation Details.}
Our training data is sourced from the 3D-Alpaca dataset~\cite{ye2025shapellm}, which comprises approximately 2.56M multimodal samples spanning text-to-3D, image-to-3D, 3D-captioning, and 3D-editing tasks. Each 3D asset is rendered from four orthogonal views and paired with GPT-generated captions, providing rich interleaved supervision across modalities. We initialize our 3D-ULM from ShapeLLM-Omni~\cite{ye2025shapellm}, which we extend with our octant-based 3D VQ-VAE for compact geometry representation and geometric CoT reasoning. Training follows the 3D Co-GRPO framework with a base learning rate of $1\times10^{-6}$ and KL regularization $\beta=0.01$. The model is trained for 40k steps with batch size 256 on 8$\times$L40 GPUs.

For reward computation, we render multi-view images for each generated mesh and evaluate them using HPS~\cite{wu2023human} for the Human Preference Critic, ShapeLLM~\cite{qi2024shapellm} for the 3D Understanding Critic, and ULIP~\cite{xue2024ulip} embeddings for the Text–3D Alignment Critic. 
Finally, the Physical Coherence Critic is implemented through geometry-based evaluation on the generated meshes.
%All critic scores are normalized to $[0,1]$ and fused as in Eq.~\ref{eq:grpo_reward} to guide policy optimization under the GRPO objective (Eq.~\ref{eq:grpo_objective}). 
More details in the Appendix.
\begin{table}[t!]
  \centering
  \caption{\textbf{Evaluation of general conversational and reasoning abilities on standard language benchmarks.} We compare \modelname against top-tier general vision-language models (VLMs) and 3D-specific language models. Our model demonstrates SoTA or competitive language understanding and reasoning performance. \colorbox{CustomLightPurple}{Best} and \colorbox{CustomLightLightPurple}{second-best} are highlighted.\looseness-1}
  \label{tab:quant_r1}
  % \vspace{-0.1cm}
  \resizebox{\columnwidth}{!}{
    \begin{tabular}{cccccc}
      \toprule
      \raisebox{1ex}{Benchmark} &
      \shortstack{Qwen2.5-\\vl-7B} &
      \shortstack{LLaMA3.2-\\Vision-11B} &
      \shortstack{LLaMA-\\Mesh-8B} &
      \shortstack{ShapeLLM-\\Omni-7B} &
      \shortstack{\modelnamegradient} \\
      \midrule
      MMLU $\uparrow$ & \colorbox{CustomLightLightPurple}{67.5} & 66.2 & 59.8 & 64.3 & \colorbox{CustomLightPurple}{67.6} \\
      PIQA $\uparrow$ & \colorbox{CustomLightPurple}{81.3} & 80.1 & 79.8 & 78.9 & \colorbox{CustomLightLightPurple}{79.4} \\
      GSM8K $\uparrow$ & 43.2 & 42.1 & 37.2 & \colorbox{CustomLightLightPurple}{55.6} & \colorbox{CustomLightPurple}{57.3} \\
      SIQA $\uparrow$ & \colorbox{CustomLightLightPurple}{41.0} & 40.6 & 40.3 & \colorbox{CustomLightPurple}{41.5} & \colorbox{CustomLightPurple}{41.5} \\
      \bottomrule
    \end{tabular}
    }
\end{table}

\begin{figure*}[!t]
\centering
    \includegraphics[width=0.99\linewidth]{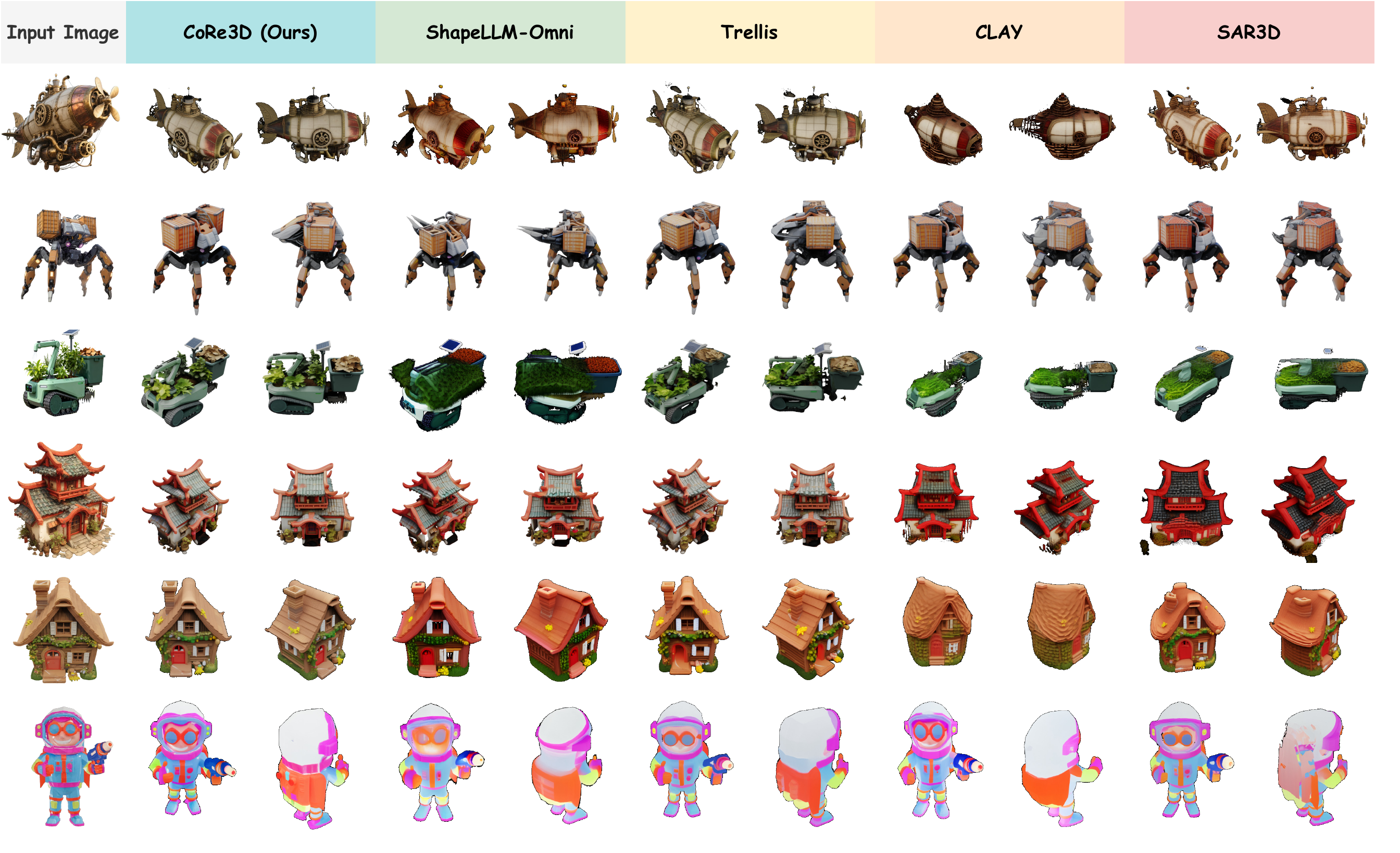}
    \caption{\textbf{Image-to-3D qualitative comparison.} Given a single input image, \modelname{} produces 3D shapes with higher geometric fidelity, cleaner topology, and stronger semantic alignment compared to baselines.\looseness-1
    }
    \label{fig:qual_results_r1}
    \vspace{-0.1cm}
\end{figure*}

\subsection{Quantitative Results}
\textbf{General Conversational Abilities.}
We first evaluate the foundational language understanding of \modelname. We compare against leading general-purpose VLMs (Qwen2.5-vl-7B~\cite{yang2024qwen2}, LLaMA3.2-Vision-11B~\cite{touvron2023llama}) and 3D-focused multimodal models (LLaMA-Mesh-8B~\cite{wang2024llama}, ShapeLLM-Omni-7B~\cite{ye2025shapellm}) on a suite of standard language benchmarks. These include MMLU~\cite{hendrycks2020measuring}, PIQA~\cite{bisk2020piqa}, GSM8K~\cite{cobbe2021training}, and SIQA~\cite{sap2019socialiqa}. As shown in \Cref{tab:quant_r1}, \modelname maintains its general conversational abilities with good language understanding and reasoning performance. These results suggest that our co-reasoning mechanism does not diminish the model’s broader reasoning capabilities but instead strengthens them.

\begin{table}[t!]
  \centering   
  \caption{\textbf{3D object captioning results on the Objaverse benchmark.} We evaluate the model's 3D caption capability. \modelname achieves state-of-the-art performance by a significant margin across all n-gram and semantic similarity metrics, demonstrating that our reasoning-driven generative training directly enhances 3D understanding. \colorbox{CustomLightPurple}{Best} and \colorbox{CustomLightLightPurple}{second-best} results are highlighted.\looseness-1}
  \label{tab:quant_r2}
  % \vspace{-0.3cm}
  \resizebox{\columnwidth}{!}{
  \begin{tabular}{l ccc cc}
      \toprule
      Model & BLEU-1 $\uparrow$ & ROUGE-L $\uparrow$ & METEOR $\uparrow$ & Sentence-BERT $\uparrow$ & SimCSE $\uparrow$ \\
      \midrule
      LLaVA-13B & 4.01 & 8.18 & 13.18 & 46.97 & 48.86 \\
      Qwen2.5-vl-7B & 4.05 & 7.85 & 14.23 & 48.90 & \colorbox{CustomLightLightPurple}{50.86} \\
      3D-LLM & 15.11 & 17.84 & 19.22 & 42.36 & 43.58 \\
      LEO & 16.98 & 20.12 & 20.91 & 48.01 & 47.25 \\
      PointLLM-13B & 3.18 & 7.54 & 12.24 & 47.89 & 49.01 \\
      ShapeLLM-Omni & \colorbox{CustomLightLightPurple}{18.92} & \colorbox{CustomLightLightPurple}{21.46} & \colorbox{CustomLightLightPurple}{22.12} & \colorbox{CustomLightLightPurple}{49.43} & 50.72 \\
      \textbf{\modelnamegradient} & \colorbox{CustomLightPurple}{24.02} & \colorbox{CustomLightPurple}{26.45} & \colorbox{CustomLightPurple}{24.98} & \colorbox{CustomLightPurple}{51.17} & \colorbox{CustomLightPurple}{52.79} \\
      \bottomrule
  \end{tabular}}
  %\vspace{-0.3cm}
\end{table}

\begin{table}[t!]
  \centering
  \caption{\textbf{Quantitative comparison of 3D generation quality for Text-to-3D and Image-to-3D tasks.} We evaluate \modelname against state-of-the-art generative models. Results show \modelname achieves competitive performance on all metrics for both tasks. \colorbox{CustomLightPurple}{Best} and \colorbox{CustomLightLightPurple}{second-best} results are highlighted.}
  \label{tab:quant_r3}
    \resizebox{\columnwidth}{!}{
    \begin{tabular}{l | ccc | ccc}
      \toprule
      \multirow{3}{*}{Method} &
      \multicolumn{3}{c|}{Text-to-3D} &
      \multicolumn{3}{c}{Image-to-3D} \\
      
      & CLIP $\uparrow$
      & FD$_\mathrm{incep}\downarrow$
      & KD$_\mathrm{incep}\downarrow$
      & CLIP $\uparrow$
      & FD$_\mathrm{incep}\downarrow$
      & KD$_\mathrm{incep}\downarrow$ \\
      \hline
      SAR3D & 0.23 & 28.4 & 0.27 & 
               0.84 & 22.1 & 0.18 \\
      CLAY & 0.27 & 23.9 & 0.21 & 
              \colorbox{CustomLightLightPurple}{0.85} & 13.5 & 0.10 \\
      Trellis & \colorbox{CustomLightLightPurple}{0.29} & \colorbox{CustomLightLightPurple}{18.6} & \colorbox{CustomLightLightPurple}{0.19} &
                 \colorbox{CustomLightLightPurple}{0.85} & \colorbox{CustomLightPurple}{10.9} & \colorbox{CustomLightPurple}{0.08} \\
      ShapeLLM-Omni & 0.27 & 24.4 & 0.24 & 
                       0.84 & 14.1 & \colorbox{CustomLightLightPurple}{0.09} \\
      \textbf{\modelnamegradient} & \colorbox{CustomLightPurple}{0.30} & \colorbox{CustomLightPurple}{18.5} & \colorbox{CustomLightPurple}{0.18} & 
                         \colorbox{CustomLightPurple}{0.86} & \colorbox{CustomLightLightPurple}{11.2} & \colorbox{CustomLightPurple}{0.08} \\
      \bottomrule
    \end{tabular}
    }
      %\vspace{-0.3cm}
\end{table}

\noindent \textbf{3D Object Understanding.}
Beyond general language abilities, a key goal of our work is to excel at 3D understanding. We test this reciprocal capability using the 3D object captioning task on an Objaverse~\cite{deitke2023objaversexl} held-out set that is isolated from our training data. We compare \modelname against prominent general-purpose VLMs (LLaVA-13B~\cite{liu2023llava}, Qwen2.5-vl-7B) and specialized 3D-language models (3D-LLM~\cite{hong20233d}, LEO~\cite{huang2024embodiedgeneralistagent3d}, PointLLM-13B~\cite{xu2024pointllm}, ShapeLLM-Omni~\cite{ye2025shapellm}). Performance is measured using n-gram matching (BLEU-1, ROUGE-L, METEOR) and semantic embedding similarity (Sentence-BERT, SimCSE) following the evaluation settings in PointLLM. The results in \Cref{tab:quant_r2} show that \modelname decisively outperforms all baselines across all five metrics. These results quantitatively confirm that our co-reasoning method substantially improves 3D understanding.

\begin{figure*}[!t]
\centering
    \includegraphics[width=0.99\linewidth]{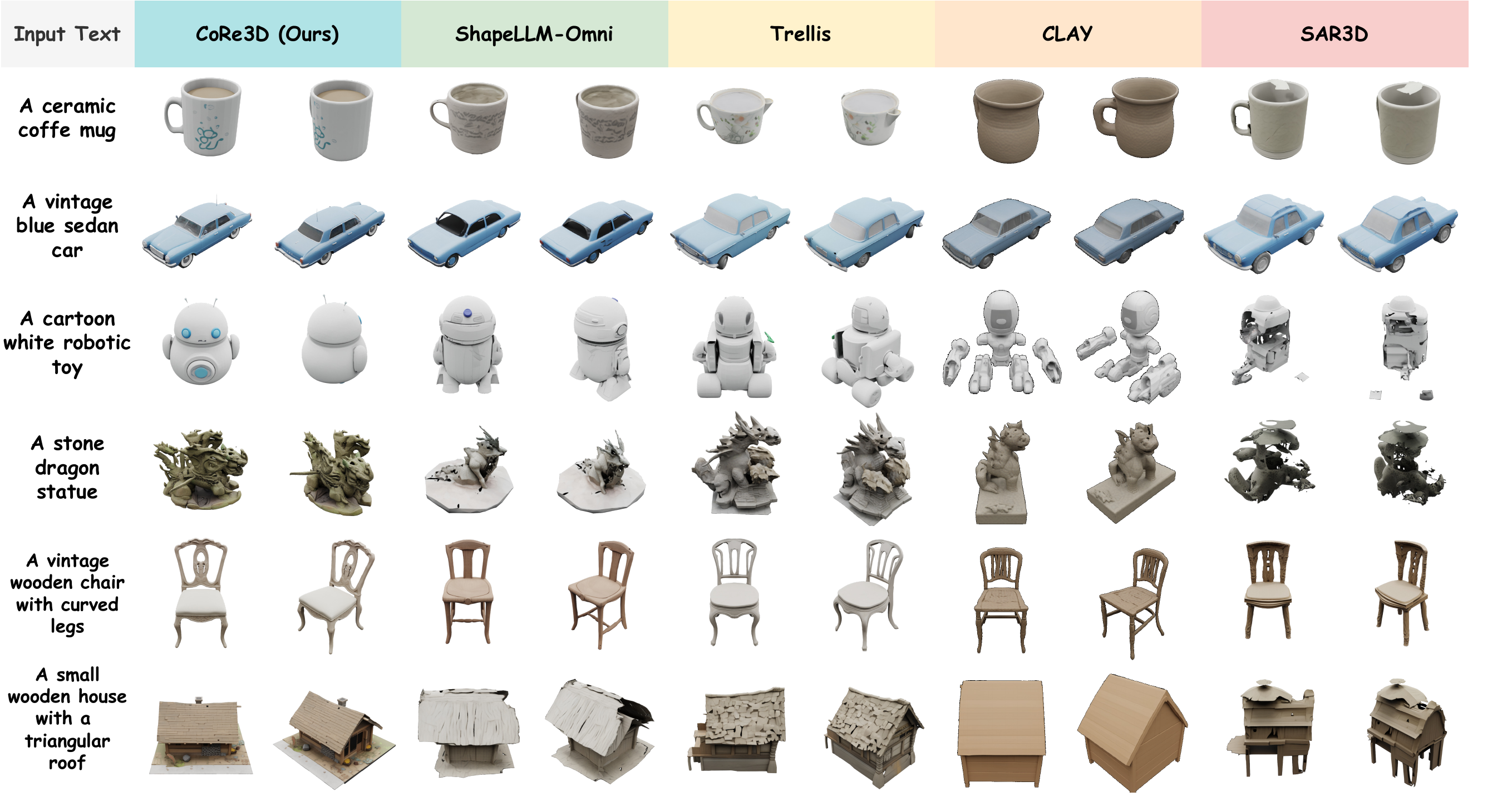}
    \vspace{-0.2cm}
    \caption{\textbf{Text-to-3D qualitative comparison.}
    \modelname generates 3D objects that more faithfully follow the textual prompt.}
    \label{fig:qual_results_r2}
    \vspace{-0.3cm}
\end{figure*}

\noindent \textbf{3D-Generation.}
We evaluate the core generative capabilities of \modelname on both Text-to-3D and Image-to-3D synthesis. We compare against several leading methods, including SAR3D~\cite{chen2024sar3d}, CLAY~\cite{zhang2024clay}, Trellis~\cite{xiang2024structured}, and ShapeLLM-Omni~\cite{ye2025shapellm}. We use three standard metrics: CLIP Score, Frechet Distance (FD), and Kernel Distance (KD) to assess the generation quality. As shown in \Cref{tab:quant_r3}, our method achieves state-of-the-art results in the Text-to-3D task, demonstrating a clear gain by ranking first across all three metrics. Our high CLIP score is a direct quantitative validation of our core contribution: the Semantic CoT, jointly optimized with GRPO and a Text-3D Alignment Critic, successfully made our model more faithful to the text. This strong performance extends to Image-to-3D generation, where our model leads in prompt alignment (CLIP 85.9) and maintains promising quality compared to the state-of-the-art 3D generation-only models.\looseness-1

\subsection{Qualitative Results}
\textbf{Standard 3D Generation.}
We first evaluate \modelname on standard image-to-3D and text-to-3D tasks, comparing it against strong baselines, including the unified 3D models ShapeLLM-Omni \cite{ye2025shapellm} and SAR3D \cite{chen2024sar3d}, as well as the generation-focused models Trellis \cite{xiang2024structured} and CLAY \cite{zhang2024clay}. As illustrated in \Cref{fig:qual_results_r1}, competing methods often exhibit misalignment with the input image or suffer from geometric artifacts. In contrast, our model generates 3D meshes with high geometric fidelity and semantic coherence, faithfully capturing the complex structures present in the source image. This superior performance extends to text-to-3D generation, as shown in \Cref{fig:qual_results_r2}. Our model achieves a more robust alignment with the text prompt. Notably, \modelname successfully interprets and renders fine-grained stylistic details, such as the ``cartoon" attribute, producing 3D meshes that align precisely with the prompt's intention.

\begin{figure*}[!t]
\centering
    \includegraphics[width=0.99\linewidth]{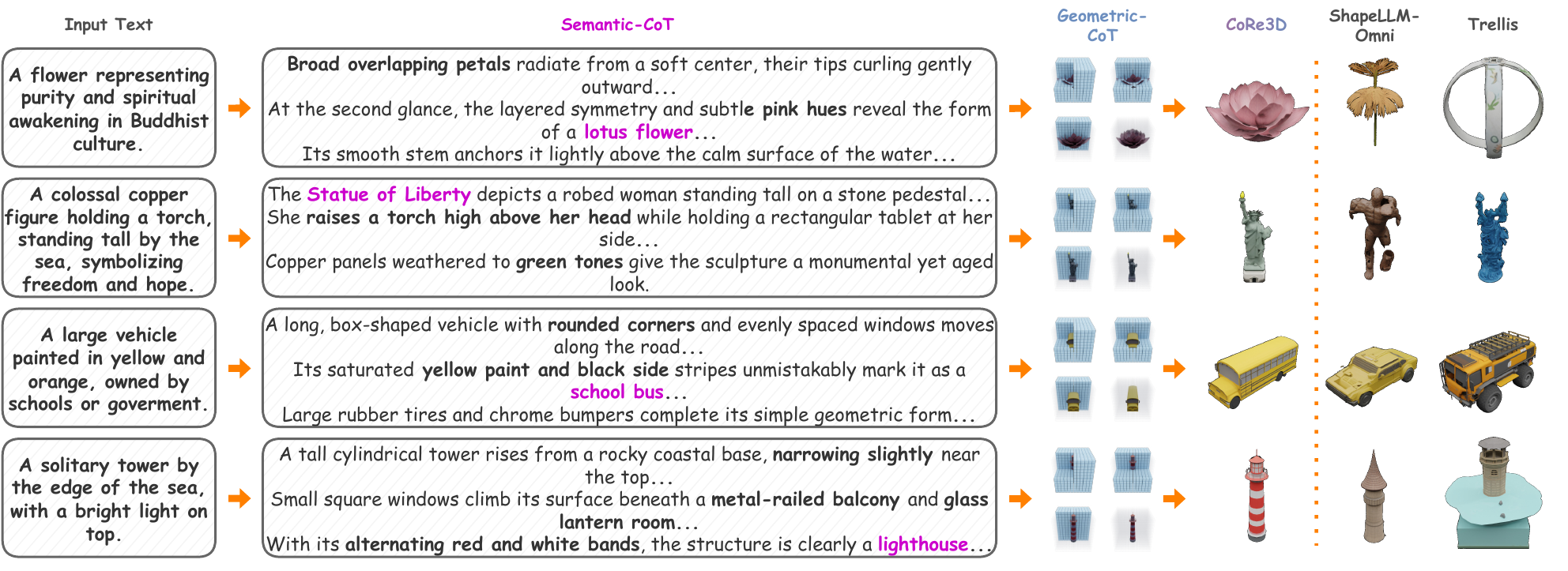}
    % \vspace{-0.3cm}
    \caption{\textbf{Qualitative results of \modelnamegradient on challenging prompts} that require inferring the correct object or interpreting implicit descriptive cues. Comparing with the base model ShapeLLM-Omni~\cite{ye2025shapellm} and state-of-the-art generation model Trellis~\cite{xiang2024structured}, our model successfully inferred the true object from the implicit prompts. (\textit{\eg ``a colossal copper figure holding a torch ... symbolizing freedom and hope''} corresponds to \textit{The Statue of Liberty.})\looseness-1}
    \label{fig:qual_r3}
    \vspace{-0.2cm}
\end{figure*}

\noindent \textbf{Reasoning-based 3D Generation.}
The robust understanding capabilities of \modelname unlock a more challenging class of 3D generation: synthesis from complex or indirect prompts. These prompts often require world knowledge and compositional reasoning to infer the user's true intent (\eg inferring ``Statue of Liberty'' from ``A colossal copper
figure holding a torch ... symbolizing freedom and hope''). 

Standard models, which lack an explicit reasoning stage, fail at this task, generating literal but semantically incorrect shapes. As demonstrated in \Cref{fig:qual_r3}, \modelname successfully navigates these challenges. Our model's Semantic-CoT first deconstructs the ambiguous prompt into a structured set of steps and effectively infers the underlying intention, allowing our model to produce 3D objects that maintain high faithfulness to the prompt's true meaning.

\begin{figure}[!t]
\centering
    \includegraphics[width=0.99\linewidth]{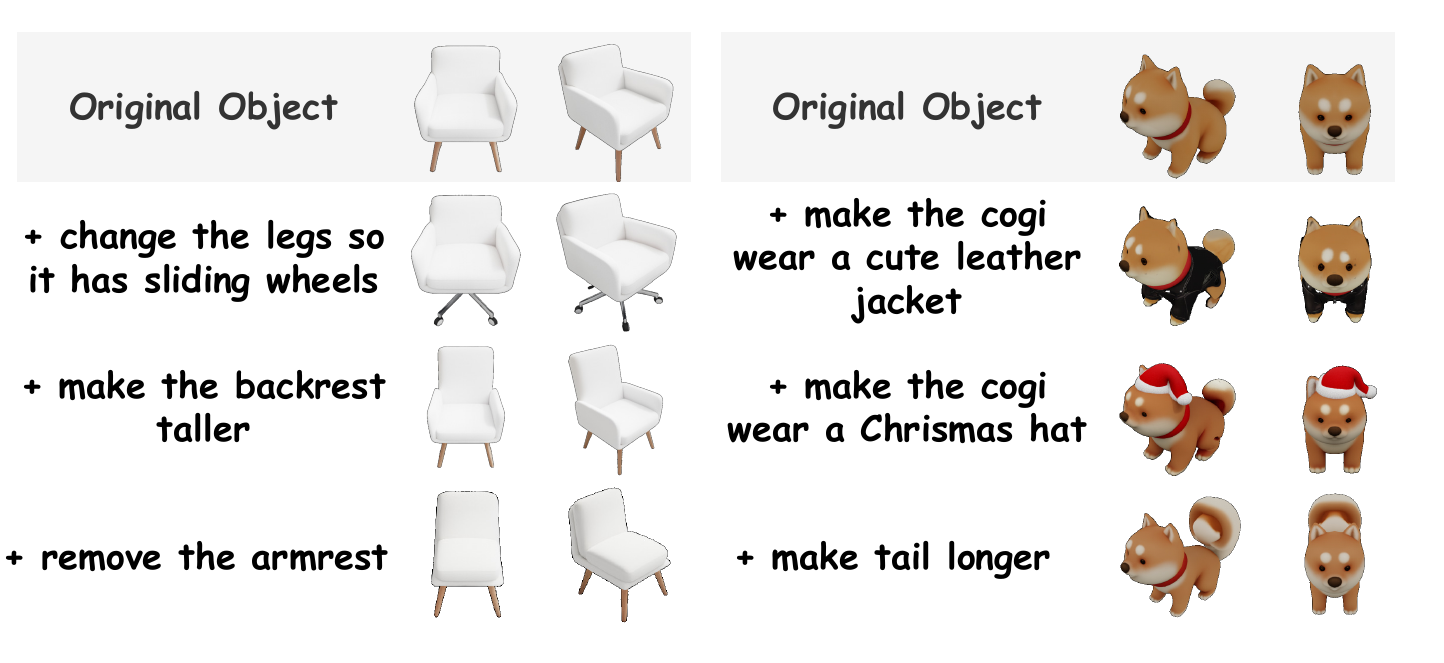}
    \caption{\textbf{Qualitative results on 3D part editing.}
    The collaborative reasoning in our framework enhances instruction comprehension, yielding edits that align more faithfully with the input text and produce 3D shapes that accurately reflect the specified modifications.\looseness-1}
    \label{fig:qual_partedit}
    \vspace{-0.3cm}
\end{figure}
\noindent \textbf{3D Part Editing.} Compared with traditional generative models, unified 3D LMs unlock a powerful language-driven paradigm for interactive 3D asset manipulation. As shown in \Cref{fig:qual_partedit}, our model can perform fine-grained part-level edits while preserving object identity and structural coherence. Additional results are provided in the Appendix.\looseness-1

\begin{figure}[!t]
\centering
    \includegraphics[width=0.98\linewidth]{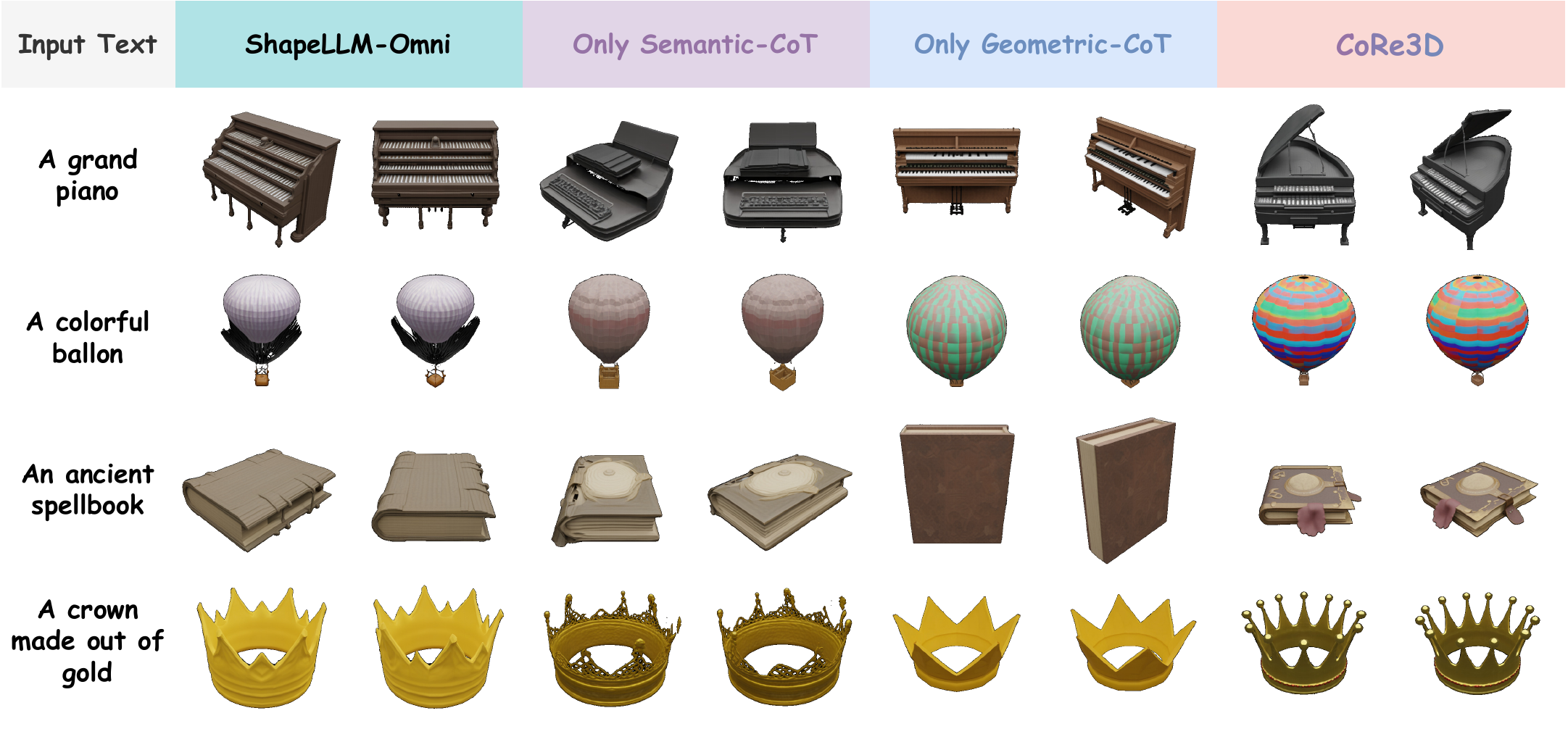}
    \caption{\textbf{Ablating Semantic CoT and Geometric CoT} compared against \modelname{} and ShapeLLM-Omni \cite{ye2025shapellm}. Results show both CoT contribute significantly towards the final performance.}
    \label{fig:ablate_r1}
\end{figure}

\subsection{Ablations}
\textbf{Semantic-CoT and Geometric-CoT.} We ablate \modelname by removing either the Semantic-CoT tokens or the Geometric-CoT tokens for the GRPO pipeline. As shown in \Cref{fig:ablate_r1}, removing Semantic-CoT leads to structurally plausible objects but lacks category- and attribute-specific cues. Conversely, removing Geometric-CoT results in objects with clear geometric distortions and simplified shapes.

\noindent \textbf{Reward Analysis}
We analyze the role of each critic and their combinations to better understand how different reward signals shape model behavior. As shown in~\Cref{tab:ablate_r1}, each critic contributes complementary improvements. Text–3D alignment yields the largest gains in caption quality, showing its importance for accurate descriptions. 3D Understanding significantly improves generation quality by enforcing stronger object-level structure. Combining multiple critics steadily improves performance: pairs such as (3DU + TA + PC) or (HP + TA + PC) achieve balanced improvements across both captioning and generation tasks. More details in the Appendix.
\begin{table}[t!]
  \centering
  \caption{\textbf{Ablation of different critics.} HP, 3DU, TA, and PC correspond to Human Preference, 3D Understanding, Text-3D Alignment, and Physical Coherence respectively. \colorbox{CustomLightPurple}{Best} and \colorbox{CustomLightLightPurple}{second-best} results are highlighted.\looseness-1}
  \label{tab:ablate_r1}
    % \vspace{-0.2cm}

\resizebox{\columnwidth}{!}{
\begin{tabular}{cccc|ccc|ccc}
  \toprule
  \multicolumn{4}{c|}{Critic} &
  \multicolumn{3}{c|}{3D Captioning} &
  \multicolumn{3}{c}{3D Generation} \\
HP & 3DU & TA & PC
    & METEOR $\uparrow$
    & Sentence\mbox{-}BERT $\uparrow$
    & SimCSE $\uparrow$
    & CLIP $\uparrow$
    & FD$_\mathrm{incep}\downarrow$
    & KD$_\mathrm{incep}\downarrow$ \\
  \midrule

\cmark & -          & -          & -          
  & 12.76 & 44.91 & 48.14 & 0.29 & 23.4 & 0.25 \\

-          & \cmark & -          & -          
  & 14.12 & 47.88 & 49.63 & 0.31 & 22.7 & 0.23 \\

-          & -          & \cmark & -          
  & 15.34 & 46.52 & 50.11 & 0.33 & 21.9 & 0.21 \\

-          & -          & -          & \cmark 
  & 13.41 & 45.76 & 48.62 & 0.30 & 20.8 & 0.22 \\

-          & \cmark & \cmark & \cmark 
  & 18.92 & 48.77 & 51.24 & 0.36 & 19.7 & \colorbox{CustomLightLightPurple}{0.19} \\

\cmark & -          & \cmark & \cmark 
  & 20.48 & 49.02 & \colorbox{CustomLightLightPurple}{51.89} & 0.37 & 19.9 & 0.20 \\

\cmark & \cmark & -          & \cmark 
  & 19.83
  & 49.88
  & 51.01
  & \colorbox{CustomLightLightPurple}{0.38}
  & 20.1
  & \colorbox{CustomLightLightPurple}{0.19} \\

\cmark & \cmark & \cmark & -          
  & \colorbox{CustomLightLightPurple}{21.55} & \colorbox{CustomLightLightPurple}{50.41} & 51.43 & 0.37 & 19.4 & 0.20 \\

\cmark & \cmark & \cmark & \cmark 
  & \colorbox{CustomLightPurple}{24.98}
  & \colorbox{CustomLightPurple}{51.17}
  & \colorbox{CustomLightPurple}{52.79}
  & \colorbox{CustomLightPurple}{0.40}
  & \colorbox{CustomLightPurple}{18.5}
  & \colorbox{CustomLightPurple}{0.18} \\

\bottomrule
\end{tabular}
}
%\vspace{-0.4cm}
\end{table}

\section{Conclusion}
We introduce \modelname, a collaborative reasoning framework that unifies semantic planning and geometric construction through a dual chain-of-thought process. By coupling these complementary reasoning levels and optimizing them jointly with Co-GRPO, our model achieves state-of-the-art performance across both 3D generation and understanding tasks. Beyond producing faithful and physically coherent 3D assets, \modelname demonstrates robust interpretive capabilities, successfully handling indirect and referring descriptions, ambiguous prompts, and fine-grained part-level edits. Our results highlight collaborative reasoning as a scalable and structure-aware foundation for general 3D intelligence.

\bibliographystyle{plainnat}
\bibliography{main}

\newpage
\appendix
\clearpage
% \setcounter{page}{1}
% \maketitlesupplementary

\section{Additional Related Work}

\noindent\textbf{Post-Pretraining Reinforcement Learning.} Since the introduction of Chain-of-Thought prompting~\cite{wei2022chain}, enhancing the reasoning ability of large language models has become an important research focus. More recently, DeepSeek-R1~\cite{guo2025deepseek} advanced this direction by proposing a rule-based reward design combined with GRPO training, encouraging models to produce explicit intermediate reasoning traces before generating final answers. This paradigm has since been extended to multimodal settings~\cite{meng2025mm,yang2025r1,r1vl,huang2025vision}, where task-specific rewards guide the learning process. These reasoning-driven methods have achieved notable gains across a variety of challenging tasks~\cite{liu2024mmbench}, including mathematical problem solving~\cite{zhang2024mathverse,zhang2024mavis} and code generation~\cite{jain2024livecodebench,liu2025purpcode}. Our proposed \modelnamegradient framework builds on this line of work by extending GRPO from text-only settings to a unified 3D understanding and 3D generation model. We employ a collaborative reasoning strategy that tightly couples linguistic semantics with 3D geometry, leading to more coherent and interpretable generation.\looseness-1

\noindent\textbf{Octant-based 3D Representations.} 
The concept of octree has been used in a wide range of 3D geometric processing applications, including point cloud compression~\cite{schnabel2006octree}, 3D texturing~\cite{benson2002octree, xiong2025texgaussian}, and multi-view scene reconstruction~\cite{szeliski1993rapid,yu2021plenoctrees}. Beyond these foundational uses, octree structures have become integral to efficient shape analysis in large-scale environments~\cite{wang2017cnn,riegler2017octnet,wang2023octformer, wu2024point, griffiths2025hotformerloc}. In the realm of generative modeling, recent advancements ~\cite{deng2025efficient, wei2025octgpt} employ adaptive tokenization or multi-scale autoregressive strategies to allocate computational resources to geometrically complex regions dynamically. Diverging from these adaptive hierarchical methods, which often result in variable-length sequences or complex tree traversals, our approach utilizes a fixed-resolution octant-based 3D VQ-VAE to discretize the 3D volume into uniform blocks. This design preserves essential spatial locality while maintaining a compact token sequence, thereby facilitating stable and interpretable geometric chain-of-thought reasoning within a unified 3D-LLM framework.\looseness-1

\section{Implementation Details}
\label{app:implement_details}

\begin{figure*}[!t]
\centering
    \includegraphics[width=0.98\textwidth]{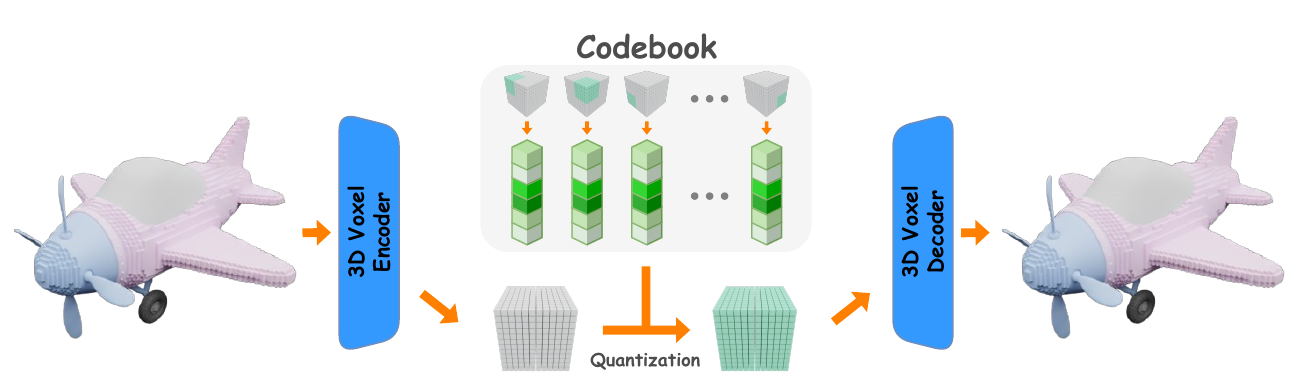}
    % \vspace{-0.3cm}
    \caption{\textbf{Overview of Octant-based 3D VQVAE.} The voxelized geometry is encoded into latent blocks, quantized using a shared codebook of 3D embeddings, and decoded back into a high-fidelity voxel grid. }
    \label{fig:vavqe}
    \vspace{-0.2cm}
\end{figure*}

\noindent\textbf{Octant-based Autoregressive Model.}
Autoregressive generation over 3D volumetric representations poses a unique challenge: the model must preserve spatial locality while maintaining a tractable token length. Traditional raster-order serialization severely disrupts locality in 3D, making next-token prediction unnecessarily difficult. Inspired by ideas from hierarchical octant-based models such as OctFormer~\cite{wang2023octformer} and OctGPT~\cite{wei2025octgpt}, we adopt an octant-structured tokenization strategy tailored to our VQ-VAE latent space. Unlike hierarchical octrees, our formulation operates on a single-scale $16^3$ latent grid while preserving the locality benefits emphasized in prior octant-based architectures.\looseness-1

As shown in \Cref{fig:vavqe}, our voxel VQ-VAE maps each $64^3$ input volume to a $16^3$ latent grid. To reduce sequence length while preserving local geometric structure, we partition this latent grid into non-overlapping $2{\times}2{\times}2$ neighborhoods. Each neighborhood forms a local \emph{octant block} that contains eight spatially adjacent latent cells. Concatenating the features of these eight cells produces a single octant token, yielding exactly $8{\times}8{\times}8 = 512$ tokens per object. These tokens summarize compact spatial regions and maintain local geometry and appearance cues.

To serialize the latent volume into an autoregressive sequence, we employ a Morton (Z-order) space-filling curve, which preserves spatial locality more effectively than raster or lexicographic scanning. Let $\mathcal{O} = [o_1, o_2, \dots, o_{512}]$
denote the sequence of octant tokens arranged in Morton order. The generative process factorizes the distribution over the latent volume as\looseness-1
\begin{equation}
    \setlength{\abovedisplayskip}{6pt}
\setlength{\belowdisplayskip}{6pt}
p(\mathcal{O}) = \prod_{i=1}^{512} p(o_i \mid o_{<i},\, \mathcal{S}_{\text{sem}}),
\end{equation}
where the semantic chain-of-thought $\mathcal{S}_{\text{sem}}$ provides high-level cues that guide geometric synthesis.

To encode spatial location, we attach a learned positional embedding to each octant token, keyed by its block index $(x_b, y_b, z_b)$ in the $8{\times}8{\times}8$ grid. This embedding is injected \emph{after} vector quantization, ensuring that the codebook remains content-centric while the autoregressive decoder remains location-aware. Together, the octant blocks and positional embeddings introduce two inductive biases: (i) each token carries high-resolution local geometric context, and (ii) the sequence ordering respects 3D locality.

During training, the semantic reasoning trace $\mathcal{S}_{\text{sem}}$ and the geometric reasoning trace $\mathcal{G}_{\text{geo}}$ jointly condition the autoregressive transformer. The semantic trace encodes global planning information such as categories and textures, while the geometric trace provides localized structural cues for neighborhood-level voxel refinement. The decoder predicts a discrete codebook index for each octant token, reconstructing the latent volume block-by-block. After predicting all 512 tokens, the VQ-VAE decoder reconstructs a dense $64^3$ voxel field, which is subsequently rendered into mesh or multi-view images. Compared to hierarchical octree models \cite{wei2025octgpt}, which must autoregressively generate thousands of binary split or leaf tokens, our design yields a significantly shorter and more expressive sequence. This compact, locality-aware autoregressive formulation is essential to our framework, enabling efficient token-level generation that tightly aligns with semantic and geometric reasoning.\looseness-1

\noindent\textbf{Physical Coherence Critic.} We use \textsc{Trimesh} to compute physical statistics and \textsc{PyMeshFix} to diagnose self-intersections. Stability is estimated by projecting the mesh center of mass onto the ground plane and checking whether this projection lies inside the convex hull of bottom support vertices extracted from the lowest-$z$ region, yielding a normalized stability score. Structural connectivity is measured by splitting the mesh into connected components and computing the fraction of faces belonging to the largest component, which penalizes fragmented or floating parts. To assess self-intersection, we run \textsc{PyMeshFix} on the original mesh and compare the number of faces before and after repair; a larger relative change indicates more severe geometric artifacts and results in a lower self-intersection score.

\noindent\textbf{Hyperparameters.} The policy is
updated using KL-regularized GRPO with ratio clipping. The KL penalty coefficient is set to $\beta$ = 0.01 and the clipping threshold to $\varepsilon$=0.1. For each prompt, we sample $K{=}4$ candidate generations and compute pairwise preferences using the four critics introduced in \Cref{sec:method_main}. The human-preference, 3D-understanding, Text-3D-alignment, and
Physical coherence critics are weighted by
$(w_H, w_V, w_X, w_P)\!=\!(0.25, 0.25, 0.25, 0.25)$.  
Optimization uses AdamW with a learning rate of $1\times 10^{-6}$, $\beta_1{=}0.9$, $\beta_2{=}0.98$, and weight decay $0.01$. We apply a 2000-step linear warmup followed by cosine decay. For stability, the policy
KL is capped at $1.2\times$ its exponential moving average, and gradients are clipped with a global norm of $1.0$.

\noindent\textbf{Evaluation Metrics.} We report common text-generation metrics, \ie, BLEU-1 \cite{papineni2002bleu}, ROUGE-L \cite{lin2004rouge}, and METEOR \cite{banerjee2005meteor}. Although standard, these metrics often favor shorter outputs and may overlook semantic fidelity. To address this, we incorporate embedding-based similarity measures, including Sentence-BERT \cite{reimers2019sentence} and SimCSE \cite{gao2021simcse}, which evaluate semantic alignment between generated captions and human references more robustly.

\begin{figure*}[!t]
\centering
    \includegraphics[width=0.92\textwidth]{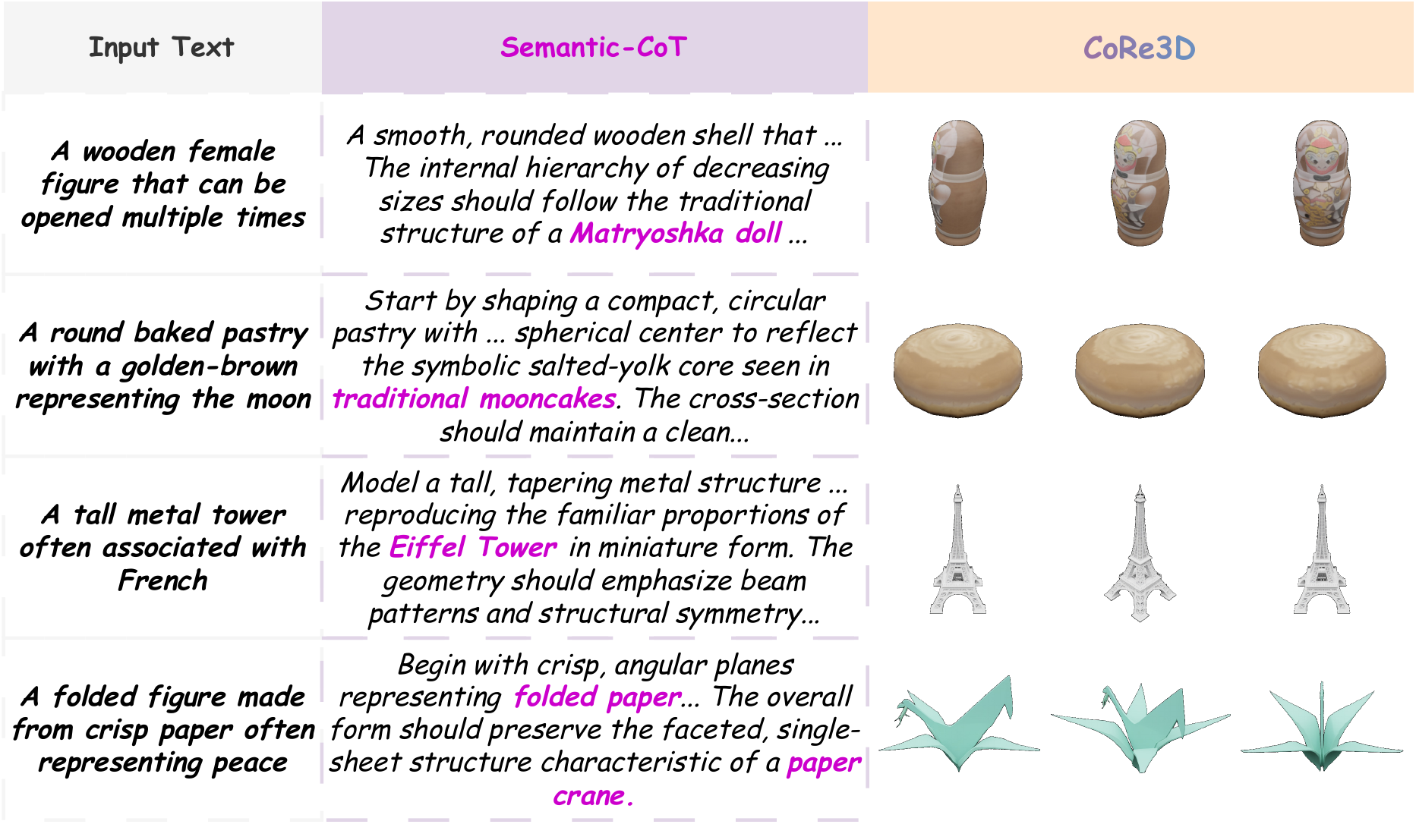}
    \vspace{-0.1cm}
    \caption{
    \textbf{Additional Results on Challenging Prompts.} \modelname successfully inferred the true object from the implicit prompts.
    }
    \label{fig:challenging_prompt_supp}
    %\vspace{-0.2cm}
\end{figure*}

\section{Additional Results}
\label{app:more_result}

\noindent\textbf{Additional Results on Challenging Prompts.} Figure~\ref{fig:challenging_prompt_supp} presents additional qualitative examples demonstrating \modelname{}’s ability to infer the correct 3D object even when the input prompt provides only indirect or symbolic descriptions. These prompts intentionally avoid naming the target object and instead describe cultural context, functional cues, or high-level visual impressions. Despite this ambiguity, \modelname consistently recovers the correct underlying structure. As shown, the model not only identifies the implicit object but also reconstructs a spatially coherent and visually faithful 3D shape. 

To our knowledge, \modelname is the first 3D generative framework capable of resolving such implicit, referential prompts through semantic reasoning, rather than relying on explicit object mentions or category supervision.\looseness-1

\begin{table*}[t!]
  \centering
  \caption{\textbf{Ablation of Semantic CoT and Geometric CoT.} \colorbox{CustomLightPurple}{Best} and \colorbox{CustomLightLightPurple}{second-best} results are highlighted.}
  \label{tab:ablate_r2}
    % \vspace{-0.2cm}
\resizebox{\linewidth}{!}{
\begin{tabular}{lcccccccc}
  \toprule
  & \multicolumn{2}{c}{\textbf{Critic}} &
  \multicolumn{3}{c}{\textbf{3D Captioning}} &
  \multicolumn{3}{c}{\textbf{3D Generation}} \\ \cmidrule{4-6}\cmidrule{7-9}
 \textbf{Model Name} & \textbf{Semantic CoT} & \textbf{Geometric CoT}
    & METEOR $\uparrow$
    & Sentence\mbox{-}BERT $\uparrow$
    & SimCSE $\uparrow$
    & CLIP $\uparrow$
    & FD$_\mathrm{incep}\downarrow$
    & KD$_\mathrm{incep}\downarrow$ \\
  \midrule

 ShapeLLM-Omni          & \xmark         & \xmark&
 12.76 & 44.91 & 48.14 & 0.29 & 23.4 & 0.25 \\

  \modelname (w/o Geometric CoT) & \cmark          & \xmark&
 \colorbox{CustomLightLightPurple}{16.42} &
 \colorbox{CustomLightPurple}{50.38} &
 \colorbox{CustomLightLightPurple}{51.14} &
 0.32 & 21.6 & 0.21 \\

 \modelname (w/o Semantic CoT) & \xmark& \cmark &
 14.89 & 47.32 & 50.41 &
 \colorbox{CustomLightLightPurple}{0.35} &
 \colorbox{CustomLightLightPurple}{19.8} &
 \colorbox{CustomLightLightPurple}{0.19} \\

 \modelnamegradient \textbf{(Ours)}         & \cmark & \cmark &
 \colorbox{CustomLightPurple}{17.03} &
 \colorbox{CustomLightLightPurple}{49.67} &
 \colorbox{CustomLightPurple}{52.11} &
 \colorbox{CustomLightPurple}{0.38} &
 \colorbox{CustomLightPurple}{18.7} &
 \colorbox{CustomLightPurple}{0.17} \\

\bottomrule
\end{tabular}
}
\vspace{-0.2cm}
\end{table*}

\noindent\textbf{Quantitative Results of Semantic-CoT and Geometric CoT Ablation.} We conduct a quantitative analysis to isolate the contributions of the two complementary reasoning CoTs. As shown in \Cref{tab:ablate_r2}, introducing \textbf{Semantic CoT} yields the largest gains in 3D captioning quality, improving METEOR, Sentence-BERT, and SimCSE scores by a significant margin over ShapeLLM-Omni. This indicates that explicit semantic reasoning helps the model better decompose text prompts into linguistically grounded object attributes. In contrast, \textbf{Geometric CoT} primarily benefits 3D generation. By encouraging structured spatial reasoning over octant-level geometry, it substantially improves CLIP similarity and reduces both FD and KD, demonstrating its effectiveness in guiding the model toward globally coherent 3D shapes. Our full model, \modelnamegradient, achieves the best overall performance across all metrics. These results verify that collaborative reasoning is essential for robust 3D understanding and generation.\looseness-1

\begin{figure*}[!t]
\centering
    \includegraphics[width=0.92\textwidth]{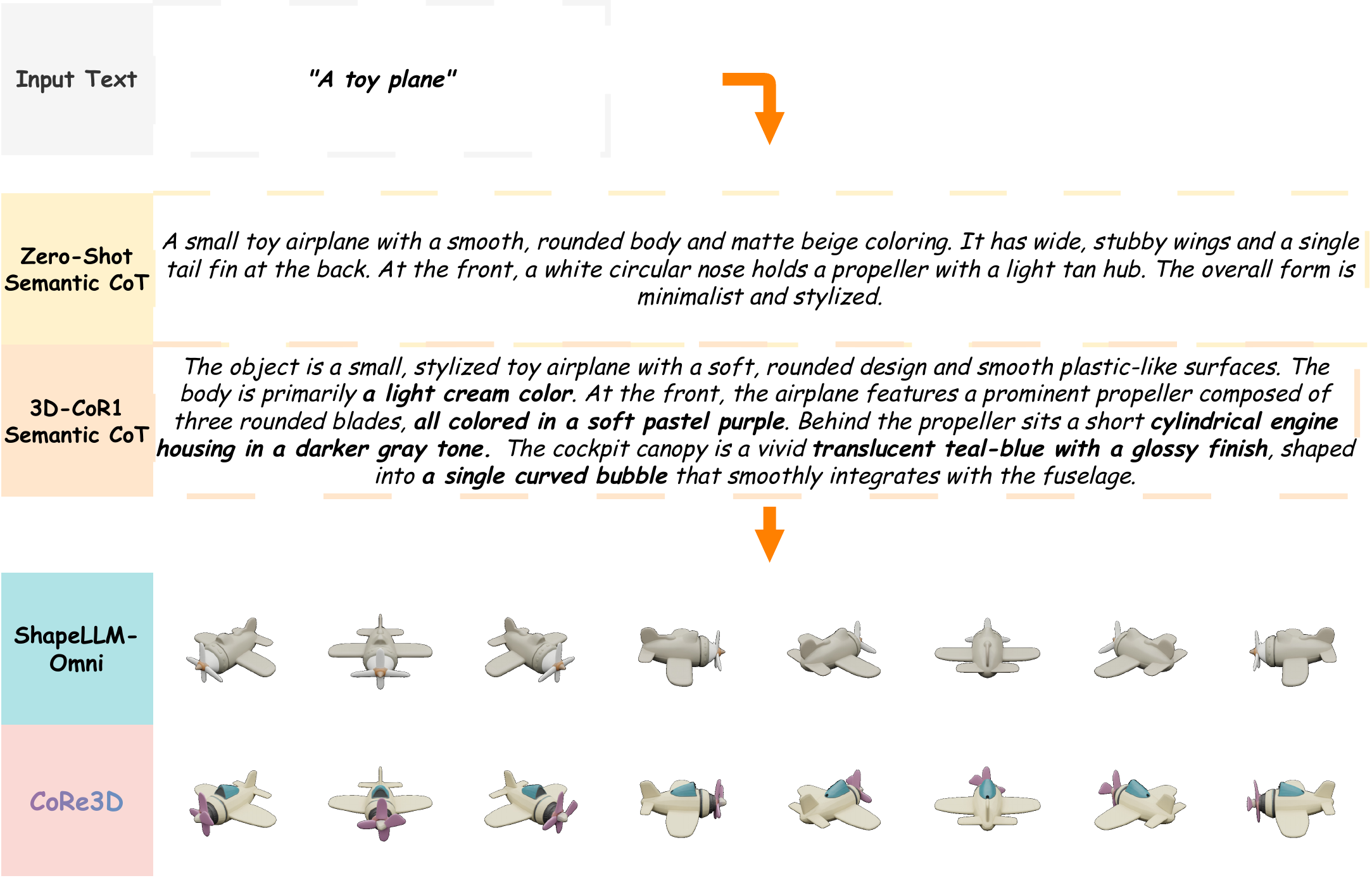}
    % \vspace{-0.3cm}
    \caption{
    \textbf{Comparison with Zero-shot CoT.} Zero-shot CoT produces shallow, generic structural descriptions, leading to oversimplified 3D shapes.
    }
    
    \label{fig:semantic_compare1}
\end{figure*}

\begin{figure*}[!t]
\centering
    \includegraphics[width=0.92\textwidth]{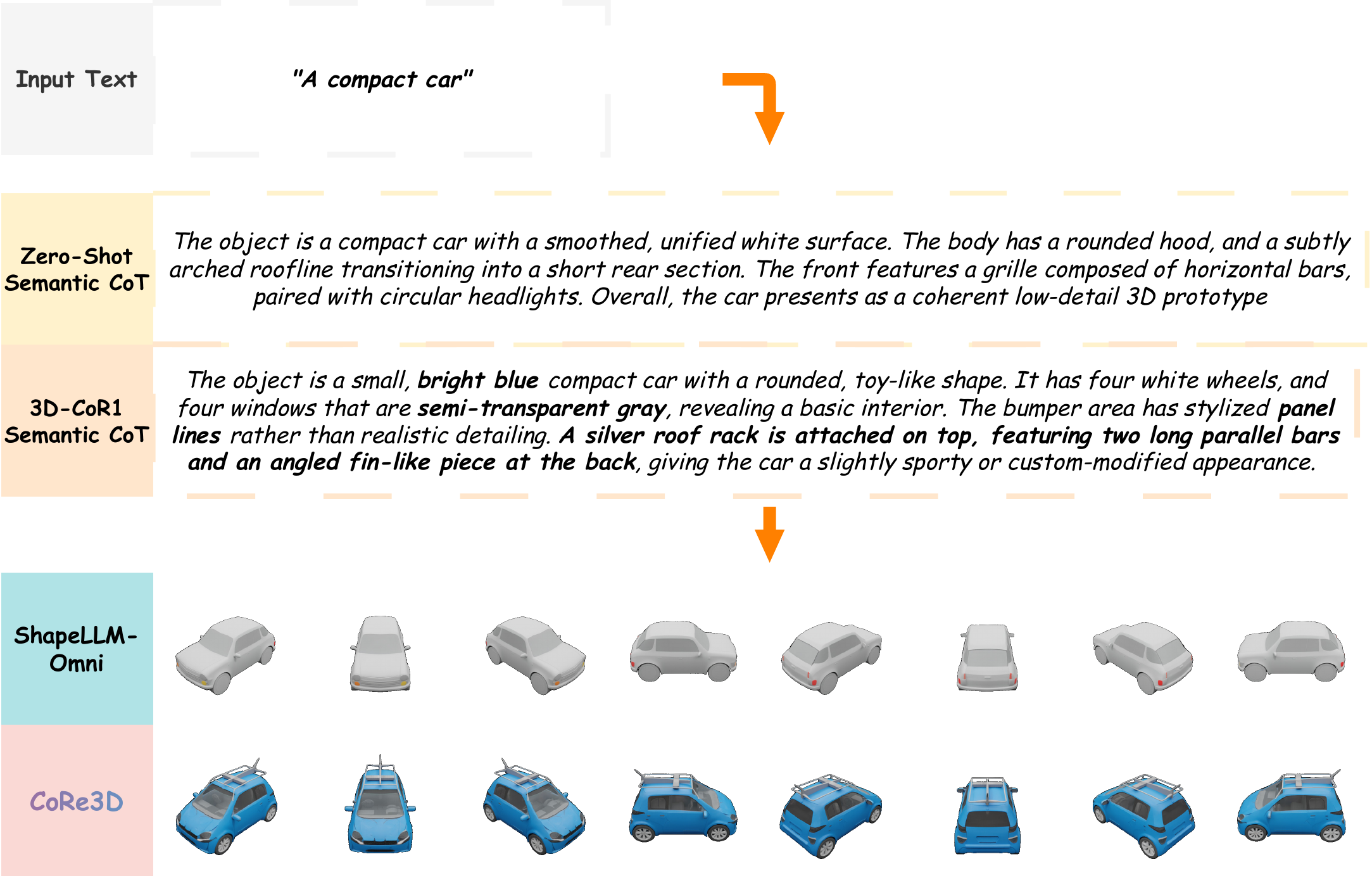}
    % \vspace{-0.3cm}
    \caption{\textbf{Comparison with Zero-shot CoT (2).} Another example showing that zero-shot CoT provides limited guidance, while our trained semantic-level CoT yields detailed structural cues.}
    \label{fig:semantic_compare2}
    % \vspace{-0.2cm}
\end{figure*}

\noindent\textbf{Comparison with Zero-shot CoT.} ~\Cref{fig:semantic_compare1} and ~\Cref{fig:semantic_compare2} compare our trained semantic-level CoT with a zero-shot CoT baseline (Qwen2.5-vl-7B~\cite{bai2025qwen25vl}). In the zero-shot setting, the model is prompted to produce a free-form structural description before generating the 3D object, but this unguided semantic reasoning provides only shallow, low-information descriptions. For instance, in the “compact car” example, the zero-shot CoT yields a generic outline of a smooth white prototype with no color, style cues, or structural modifiers, leading to an oversimplified reconstruction.  In contrast, the \modelname semantic-level CoT produces richer and more actionable structural cues, such as bright blue body color, semi-transparent windows, stylized panel lines, or the presence of a roof rack with parallel bars and a fin-like rear element (bottom CoT row), and the model correspondingly reconstructs a much more faithful 3D shape. These results show that zero-shot CoT lacks the necessary semantic grounding for 3D generation, whereas \modelname learns to produce CoT that is both structurally informative and tightly aligned with the generation process. This highlights the necessity of our collaborative reasoning pipeline.\looseness-1

\begin{figure*}[!t]
\centering
    \includegraphics[width=0.97\textwidth]{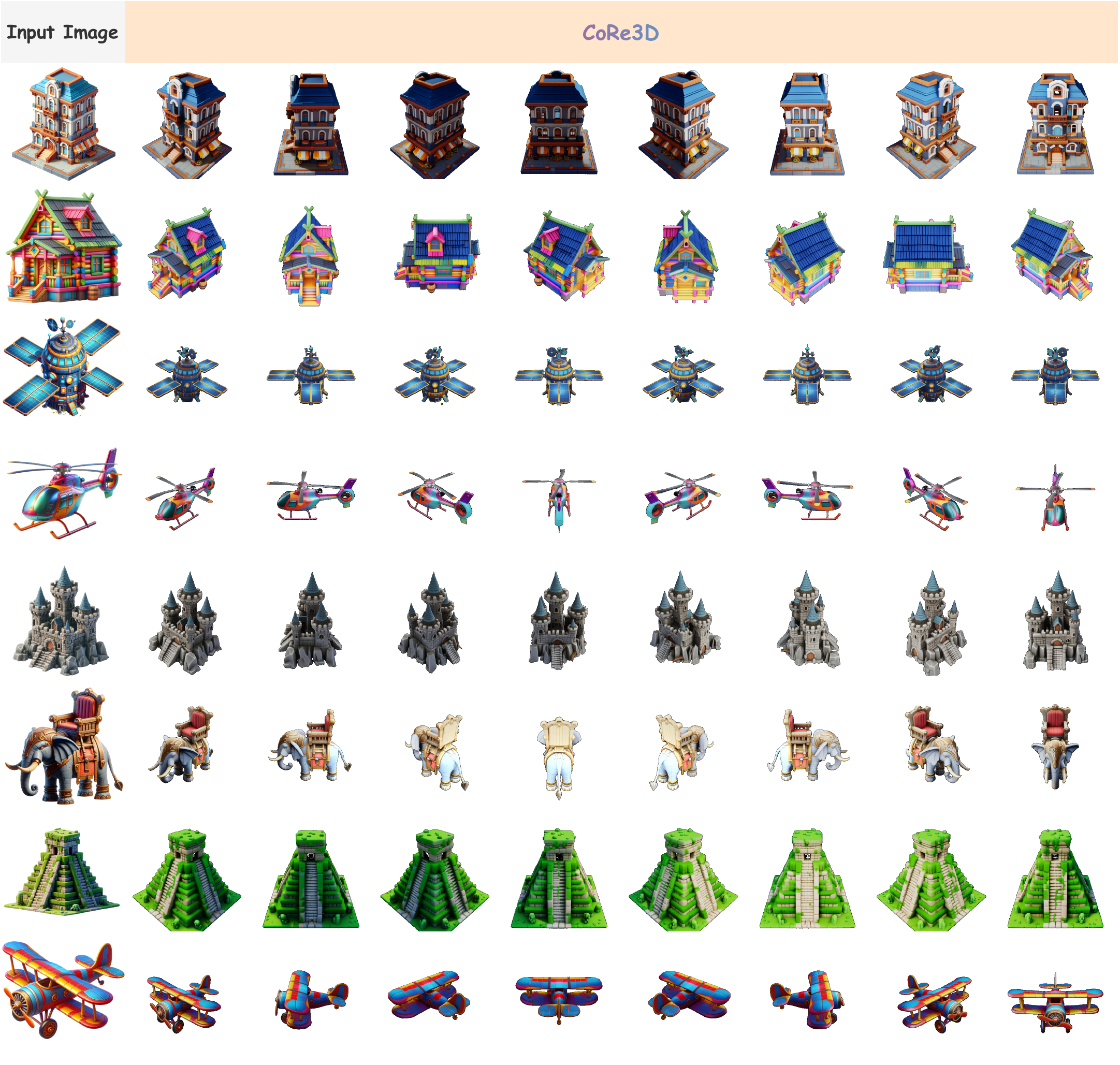}
    \vspace{-0.3cm}
    \caption{\textbf{Additional Image-to-3D Results.} We use visually complex image prompts to demonstrate that \modelname achieves strong 3D reconstruction capability, producing coherent geometry and spatially consistent shapes even for challenging inputs.}
    \label{fig:additional_image3d}
\end{figure*}

\noindent\textbf{Additional Image-to-3D Results.} Figure~\ref{fig:additional_image3d} shows additional qualitative results from our image-to-3D pipeline. Despite being a unified model rather than a specialized reconstruction system, \modelname generates stable and coherent 3D shapes even from visually complex or cluttered inputs. The model maintains globally consistent geometry, plausible spatial structure, and strong color fidelity, delivering reconstructions that capture the essential form, material cues, and overall visual identity of the input objects.

\begin{table}[t!]
  \centering
  \caption{\textbf{Ablation of Octant Depth in the Octant-based 3D VQ-VAE.}
  Increasing depth increases spatial granularity (8 $\rightarrow$ 4096 octants) and improves fidelity up to Depth 3, after which autoregressive instability degrades performance. 
  \colorbox{CustomLightPurple}{Best} and \colorbox{CustomLightLightPurple}{second-best}.}
  \label{tab:ablate_r3}
  % \vspace{-0.3cm}
\resizebox{0.99\columnwidth}{!}{
\begin{tabular}{cccccccc}
  \toprule
  \multicolumn{2}{c}{\textbf{Architecture}} &
  \multicolumn{3}{c}{\textbf{3D Captioning}} &
  \multicolumn{3}{c}{\textbf{3D Generation}} \\ \cmidrule{3-5} \cmidrule{6-8}
    \textbf{Depth} & \textbf{\# Octants}
    & METEOR $\uparrow$
    & SBERT $\uparrow$
    & SimCSE $\uparrow$
    & CLIP $\uparrow$
    & FD$_\mathrm{incep}\downarrow$
    & KD$_\mathrm{incep}\downarrow$ \\
  \midrule

  \textbf{1} & 8 &
  7.34 &
  41.82 &
  43.10 &
  0.21 &
  42.3 &
  0.34 \\

  \textbf{2} & 64 &
  12.87 &
  46.51 &
  47.88 &
  0.29 &
  29.4 &
  0.25 \\

  \textbf{3 (Ours)} & 512 &
  \colorbox{CustomLightPurple}{17.03} &
  \colorbox{CustomLightLightPurple}{49.67} &
  \colorbox{CustomLightPurple}{52.11} &
  \colorbox{CustomLightPurple}{0.38} &
  \colorbox{CustomLightPurple}{18.7} &
  \colorbox{CustomLightPurple}{0.17} \\

  \textbf{4} & 4096 &
  \colorbox{CustomLightLightPurple}{16.44} &
  \colorbox{CustomLightPurple}{49.92} &
  51.36 &
  \colorbox{CustomLightLightPurple}{0.37} &
  \colorbox{CustomLightLightPurple}{19.0} &
  \colorbox{CustomLightLightPurple}{0.18} \\
  \bottomrule
\end{tabular}
}
\vspace{-0.3cm}
\end{table}

\noindent\textbf{Optimal Octant Layers.} Table~\ref{tab:ablate_r3} evaluates how the choice of octant depth and therefore the total number of octant tokens affects the performance of our octant-based 3D VQ-VAE.  Increasing the depth refines the spatial partitioning of the latent volume (from $8$ to $4096$ octants), granting the model access to progressively finer geometric detail. We observe a consistent upward
trend from Depth~1 to Depth~3 across all captioning and generation metrics. Depth~1 (only $8$ octants) severely under-parameterizes local geometry, leading to weak semantic alignment and the worst FD/KD scores. Depth~2 (64 octants) recovers coarse global structure but still lacks local detail, resulting in moderate improvements. Depth~3 provides the best balance between spatial granularity and autoregressive stability. It achieves the highest scores for METEOR, SimCSE, CLIP, and the lowest FD/KD, demonstrating that this level of decomposition offers enough local resolution to capture high-frequency geometry without excessively
lengthening the autoregressive sequence. We use Depth~3 as our default. At Depth~4, although the finer subdivision (4096 octants) slightly
improves SBERT similarity and CLIP alignment, overall performance begins to degrade.\looseness-1
\begin{table}[t!]
  \centering
  \caption{\textbf{Ablation of Codebook Size in the Octant-based 3D VQ-VAE.}
  Larger codebooks reduce quantization error but exhibit diminishing returns
  and instability at extreme scales. Octant depth is fixed at 512 octants (Depth~3).
  \colorbox{CustomLightPurple}{Best} and 
  \colorbox{CustomLightLightPurple}{second-best}.}
  \label{tab:ablate_r4}
  % \vspace{-0.3cm}

\resizebox{\columnwidth}{!}{
\begin{tabular}{cccccccc}
  \toprule
  \multicolumn{2}{c}{\textbf{Architecture}} &
  \multicolumn{3}{c}{\textbf{3D Captioning}} &
  \multicolumn{3}{c}{\textbf{3D Generation}} \\ \cmidrule{3-5} \cmidrule{6-8}
    \textbf{Codebook Size} & \textbf{Octants}
    & METEOR $\uparrow$
    & SBERT $\uparrow$
    & SimCSE $\uparrow$
    & CLIP $\uparrow$
    & FD$_\mathrm{incep}\downarrow$
    & KD$_\mathrm{incep}\downarrow$ \\
  \midrule

  \textbf{2048} & 512 &
  13.81 &
  46.92 &
  47.44 &
  0.31 &
  27.8 &
  0.24 \\

  \textbf{4096} & 512 &
  15.92 &
  48.51 &
  50.03 &
  0.34 &
  22.1 &
  0.20 \\

  \textbf{8192 (Ours)} & 512 &
  \colorbox{CustomLightPurple}{17.03} &
  \colorbox{CustomLightLightPurple}{49.67} &
  \colorbox{CustomLightPurple}{52.11} &
  \colorbox{CustomLightPurple}{0.38} &
  \colorbox{CustomLightPurple}{18.7} &
  \colorbox{CustomLightPurple}{0.17} \\

  \textbf{16384} & 512 &
  \colorbox{CustomLightLightPurple}{16.77} &
  \colorbox{CustomLightPurple}{49.98} &
  \colorbox{CustomLightLightPurple}{51.62} &
  \colorbox{CustomLightLightPurple}{0.37} &
  \colorbox{CustomLightLightPurple}{19.3} &
  \colorbox{CustomLightLightPurple}{0.18} \\
  \bottomrule
\end{tabular}
}
\vspace{-0.3cm}
\end{table}

\noindent \textbf{Optimal Codebook Size.} Table~\ref{tab:ablate_r4} evaluates how the capacity of the VQ-VAE codebook affects both semantic alignment and 3D generation fidelity. A small codebook (2048 entries) creates a severe quantization bottleneck. This under-expressiveness leads to noticeably lower METEOR, SBERT, and SimCSE scores, along with substantially worse FD/KD metrics. Increasing the codebook to 4096 entries alleviates this issue and yields consistent
gains across all evaluation dimensions. Our default setting with an 8192-entry codebook achieves the best overall performance. Expanding the codebook further to 16384 entries results in only marginal improvements on SBERT and CLIP similarity, but degrades generation fidelity. These findings indicate that our 8192-entry codebook is optimal.\looseness-1

\section{Broader Impacts}
\modelname advances unified 3D intelligence by enabling models to both \emph{interpret} and \emph{construct} 3D objects through collaborative semantic- geometric- reasoning. Such capabilities can benefit a wide spectrum of applications. In robotics and embodied AI, improved spatial reasoning may support better manipulation, affordance understanding, and task planning. In simulation and digital twin systems, controllable 3D generation can streamline asset creation and accelerate workflows in engineering and education. 

At the same time, such frameworks carry several potential risks. Unified 3D models may unintentionally reproduce private or proprietary content, and generative pipelines could be misused to create realistic yet deceptive 3D assets. 
\modelname is intended for research and educational use, and we encourage responsible deployment that emphasizes transparency, provenance tracking, consent for training data, and adherence to domain-specific usage guidelines. With such measures in place, we believe the benefits of collaborative reasoning for 3D understanding and generation outweigh the potential risks. 

\clearpage

\end{document}